\documentclass[10pt,twocolumn,letterpaper]{article}
\usepackage{iccv}
\usepackage{times}
\usepackage{epsfig}
\usepackage{graphicx}
\usepackage{amsmath}
\usepackage{amssymb}
\usepackage{caption}
\usepackage{subcaption}
\usepackage{booktabs}
\usepackage{bbm}

\usepackage[pagebackref=true,breaklinks=true,letterpaper=true,colorlinks,bookmarks=false]{hyperref}

\iccvfinalcopy 

\def\iccvPaperID{1766} 

\ificcvfinal\fi

\begin{document}

\title{End-to-End Urban Driving by Imitating a Reinforcement Learning Coach}

\author{Zhejun Zhang\textsuperscript{1}, Alexander Liniger\textsuperscript{1}, Dengxin Dai\textsuperscript{1,2}, Fisher Yu\textsuperscript{1} and Luc Van Gool\textsuperscript{1,3}\\
\textsuperscript{1}Computer Vision Lab, ETH Z\"urich,
\textsuperscript{2}MPI for Informatics, \textsuperscript{3}PSI, KU Leuven\\
{\tt\small \{zhejun.zhang,alex.liniger,dai,vangool\}@vision.ee.ethz.ch, i@yf.io}
}

\maketitle
\ificcvfinal\thispagestyle{empty}\fi

\begin{abstract}
End-to-end approaches to autonomous driving commonly rely on expert demonstrations. 
Although humans are good drivers, they are not good coaches for end-to-end algorithms that demand dense on-policy supervision. 
On the contrary, automated experts that leverage privileged information can efficiently generate large scale on-policy and off-policy demonstrations. 
However, existing automated experts for urban driving make heavy use of hand-crafted rules and perform suboptimally even on driving simulators, where ground-truth information is available. 
To address these issues, we train a reinforcement learning expert that maps bird's-eye view images to continuous low-level actions. 
While setting a new performance upper-bound on CARLA, our expert is also a better coach that provides informative supervision signals for imitation learning agents to learn from.
Supervised by our reinforcement learning coach, a baseline end-to-end agent with monocular camera-input achieves expert-level performance. Our end-to-end agent achieves a 78\% success rate while generalizing to a new town and new weather on the NoCrash-dense benchmark and state-of-the-art performance on the challenging public routes of the CARLA LeaderBoard.
\end{abstract}

\section{Introduction}
\begin{figure}[t]
\begin{center}
    \includegraphics[width=0.98\linewidth]{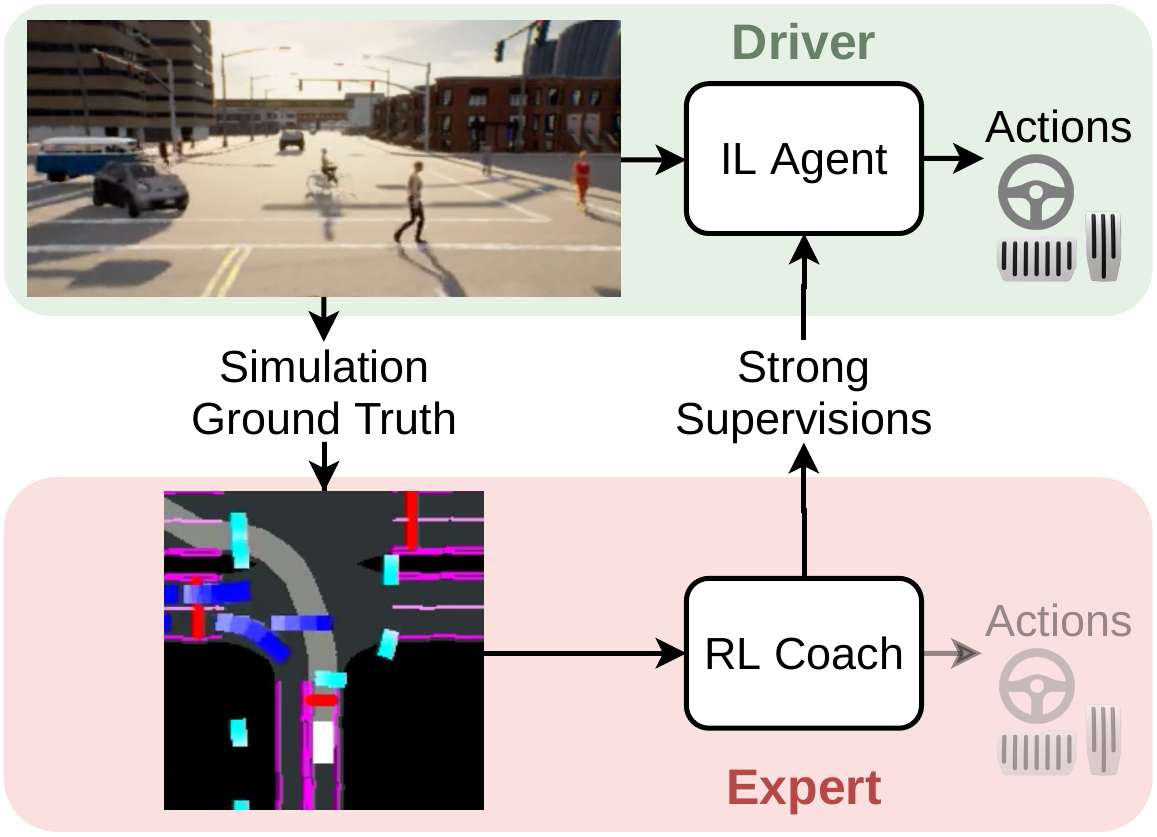}
\end{center}
\vspace{-2.5ex}
\caption{\textbf{Roach: RL coach} allows IL agents to benefit from dense and informative on-policy supervisions.}
\vspace{-2.0ex}
\label{fig:intro}
\end{figure}

Even though nowadays, most autonomous driving (AD) stacks  \cite{montemerlo2008junior,urmson2008autonomous} use individual modules for perception, planning and control, end-to-end approaches have been proposed since the 80's \cite{Pomerleau-1989-15721} and the success of deep learning brought them back into the research spotlight \cite{bojarski2016end,xu2017end}. 
Numerous works have studied different network architectures for this task \cite{DBLP:conf/rss/BansalKO19,hecker2020learning,zeng2019end}, yet most of these approaches use supervised learning with expert demonstrations, which is known to suffer from covariate shift \cite{prakash2020exploring,ross2011reduction}. 
While data augmentation based on view synthesis \cite{amini2020learning,bojarski2016end,Pomerleau-1989-15721} can partially alleviate this issue, in this paper, we tackle the problem from the perspective of expert demonstrations.

Expert demonstrations are critical for end-to-end AD algorithms.
While imitation learning (IL) methods directly mimic the experts' behavior \cite{DBLP:conf/rss/BansalKO19,codevilla2018end}, reinforcement learning (RL) methods often use expert demonstrations to improve sample efficiency by pre-training part of the model via supervised learning \cite{liang2018cirl,toromanoff2020end}.
In general, expert demonstrations can be divided into two categories: 
\emph{(i) Off-policy}, where the expert directly controls the system, and the state/observation distribution follows the expert. 
Off-policy data for AD includes, for example, public driving datasets \cite{nuscenes2019,lyft2020,yu2020bdd100k}.
\emph{(ii) On-policy}, where the system is controlled by the desired agent and the expert ``labels" the data. 
In this case, the state/observation distribution follows the agent, but expert demonstrations are accessible. 
On-policy data is fundamental to alleviate covariate shift as it allows the agent to learn from its own mistakes, which the expert in the off-policy data does not exhibit.
However, collecting adequate on-policy demonstrations from humans is non-trivial. 
While trajectories and actions taken by the human expert can be directly recorded during off-policy data collection, labeling these targets given sensor measurements turns out to be a challenging task for humans. 
In practice, only sparse events like human interventions are recorded, which, due to the limited information it contains, is hard to use for training and better suited for RL \cite{amini2020learning,kahn2021land,kendall2019learning} than for IL methods.

In this work we focus on automated experts, which in contrast to human experts can generate large-scale datasets with dense labels regardless of whether they are on-policy or off-policy.
To achieve expert-level performance, automated experts may rely on exhaustive computations, expensive sensors or even ground truth information, so it is undesirable to deploy them directly. 
Even though some IL methods do not require on-policy labeling, such as GAIL \cite{gail} and inverse RL \cite{abbeel2004apprenticeship}, these methods are not efficient in terms of on-policy interactions with the environment.

On the contrary, automated experts can reduce the expensive on-policy interactions.
This allows IL to successfully apply automated experts to different aspects of AD. 
As a real-world example, Pan et al. \cite{DBLP:conf/rss/PanCSLYTB18} demonstrated end-to-end off-road racing with a monocular camera by imitating a model predictive control expert with access to expensive sensors. In the context of urban driving, \cite{prakash2020exploring} showed that a similar concept can be applied to the driving simulator CARLA \cite{Dosovitskiy17}. 
Driving simulators are an ideal proving ground for such approaches since they are inherently safe and can provide ground truth states. 
However, there are two caveats. The first regards the ``expert" in CARLA, commonly referred to as the Autopilot (or the roaming agent). 
The Autopilot has access to ground truth simulation states, but due to the use of hand-crafted rules, its driving skills are not comparable to a human expert's. 
Secondly, the supervision offered by most automated experts is not informative. 
In fact, the IL problem can be seen as a knowledge transfer problem and just learning from expert actions is inefficient. 

To tackle both drawbacks and motivated by the success of model-free RL in Atari games \cite{hessel2018rainbow} and continuous control \cite{haarnoja2018soft}, we propose Roach (RL coach), an RL expert that maps bird's-eye view (BEV) images to continuous actions (Fig. \ref{fig:intro} bottom).
After training from scratch for 10M steps, Roach sets the new performance upper-bound on CARLA by outperforming the Autopilot. 
We then train IL agents and investigate more effective training techniques when learning from our Roach expert.
Given that Roach uses a neural network policy, it serves as a better coach for IL agents also based on neural networks.
Roach offers numerous informative targets for IL agents to learn from, which go far beyond deterministic action provided by other experts. 
Here we demonstrate the effectiveness of using action distributions, value estimations and latent features as supervisions.

Fig.~\ref{fig:intro} shows the scheme of learning from on-policy supervisions labeled by Roach on CARLA. 
We also record off-policy data from Roach by using its output to drive the vehicle on CARLA.
Leveraging 3D detection algorithms \cite{liang2019multi,wang2019monocular} and extra sensors to synthesize the BEV, Roach could also address the scarcity of on-policy supervisions in the real world.
This is feasible because on the one hand, BEV as a strong abstraction reduces the sim-to-real gap \cite{pmlr-v87-mueller18a}, and on the other hand, on-policy labeling does not have to happen in real-time or even onboard. Hence 3D detection becomes easier given the complete sequences \cite{qi2021offboard}.

In summary, this paper presents Roach, an RL expert that sets a new performance upper-bound on CARLA. Moreover, we demonstrate the state-of-the-art performance on both the NoCrash benchmark and the public routes of CARLA LeaderBoard using a single camera based end-to-end IL agent, which is supervised by Roach using our improved training scheme. 
Our repository is available at \url{https://github.com/zhejz/carla-roach}

\section{Related Work}

Since our methods are trained and evaluated on CARLA, we mainly focus on related works also done on CARLA.

\vspace{1ex}\noindent{\bf End-to-End IL:}
Dosovitskiy et al. \cite{Dosovitskiy17} introduced the CARLA driving simulator and demonstrated that a baseline end-to-end IL method with single camera input can achieve a performance comparable to a modular pipeline.
After that, CIL \cite{codevilla2018end} and CILRS \cite{codevilla2019exploring} addressed directional multi-modality in AD by using branched action heads where the branch is selected by a high-level directional command.
While the aforementioned methods are trained via behavior cloning, DA-RB \cite{prakash2020exploring} applied DAGGER \cite{ross2011reduction} with critical state sampling to CILRS.
Most recently, LSD \cite{ohn2020learning} increased the model capacity of CILRS by learning a mixture of experts and refining the mixture coefficients using evolutionary optimization. 
Here, we use DA-RB as the baseline IL agent to be supervised by Roach.

\vspace{1ex}\noindent{\bf Mid-to-X IL:}
Directly mapping camera images to low-level actions requires a large amount of data, especially if one wants generalization to diverse weather conditions. 
Mid-to-X approaches alleviate this issue by using more structured intermediate representation as input and/or output.
CILRS with coarse segmentation masks as input was studied in \cite{Behl2020IROS}. 
CAL \cite{sauer2018conditional} combines CIL and direct perception \cite{chen2015deepdriving} by mapping camera images to driving affordances which can be directly used by a rule-based low-level controller. 
LBC \cite{chen2020learning} maps camera images to waypoints by mimicking a privileged mid-to-mid IL agent similar to Chauffeurnet \cite{DBLP:conf/rss/BansalKO19}, which takes BEV as input and outputs future waypoints. 
Similarly, SAM \cite{zhao2020sam} trained a visuomotor agent by imitating a privileged CILRS agent that takes segmentation and affordances as inputs. 
Our Roach adopts BEV as the input representation and predicts continuous low-level actions.

\vspace{1ex}\noindent{\bf RL:}
As the first RL agent on CARLA, an A3C agent \cite{mnih2016asynchronous} was demonstrated in \cite{Dosovitskiy17}, yet its performance is lower than that of other methods presented in the same paper.
CIRL \cite{liang2018cirl} proposed an end-to-end DDPG \cite{LillicrapHPHETS15} agent with its actor network pre-trained via behavior cloning to accelerate online training. 
To reduce the problem complexity, Chen et al. \cite{chen2019model} investigated DDQN \cite{ddqn}, TD3 \cite{fujimoto2018addressing} and SAC \cite{haarnoja2018soft} using BEV as an input and pre-trained the image encoder with a variational auto-encoder \cite{kingma2013auto} on expert trajectories. 
State-of-the-art performance is achieved in \cite{toromanoff2020end} using Rainbow-IQN \cite{toromanoff2019deep}.
To reduce the number of trainable parameters during online training, its image encoder is pre-trained to predict segmentation and affordances on an off-policy dataset. 
IL was combined with RL in \cite{Rhinehart2020Deep} and multi-agent RL on CARLA was discussed in \cite{palanisamy2019multiagent}.
In contrast to these RL methods, Roach achieves high sample efficiency without using any expert demonstrations.

\vspace{1ex}\noindent{\bf IL with Automated Experts:}
The effectiveness of automated experts was demonstrated in \cite{DBLP:conf/rss/PanCSLYTB18} for real-world off-road racing, where a visuomotor agent is trained by imitating on-policy actions labeled by a model predictive control expert equipped with expensive sensors. 
Although CARLA already comes with the Autopilot, it is still beneficial to train a proxy expert based on deep neural networks, as shown by LBC \cite{chen2020learning} and SAM \cite{zhao2020sam}. 
Through a proxy expert, the complex to solve end-to-end problem is decomposed into two simpler stages.
At the first stage, training the proxy expert is made easier by formulating a mid-to-X IL problem that separates perception from planning. 
At the second stage, the end-to-end IL agent can learn more effectively from the proxy expert given the informative targets it supplies.
To provide strong supervision signals, LBC queries all branches of the proxy expert and backpropagates all branches of the IL agent given one data sample, whereas SAM matches latent features of the proxy expert and the end-to-end IL agent.
While the proxy expert addresses planning, it is also possible to address perception at the first stage as shown by FM-Net \cite{hou2019learning}. 
Overall, two-stage approaches achieve better performance than direct IL, but using proxy experts inevitably lowers the performance upper-bound as a proxy expert trained via IL cannot outperform the expert it imitates.
This is not a problem for Roach, which is trained via RL and outperforms the Autopilot.

\begin{figure*}[t]
     \centering
     \begin{subfigure}[b]{0.15\textwidth}
         \centering
         \includegraphics[width=\textwidth]{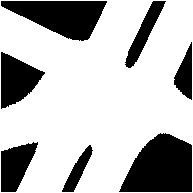}
         \caption{Drivable areas}
         \label{fig:bev_road}
     \end{subfigure}
     \hfil
     \begin{subfigure}[b]{0.15\textwidth}
         \centering
         \includegraphics[width=\textwidth]{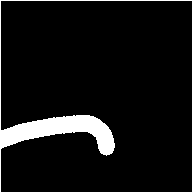}
         \caption{Desired route}
         \label{fig:bev_route}
     \end{subfigure}
     \hfil
     \begin{subfigure}[b]{0.15\textwidth}
         \centering
         \includegraphics[width=\textwidth]{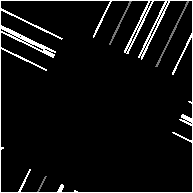}
         \caption{Lane boundaries}
         \label{fig:bev_lane}
     \end{subfigure}
     \hfil
     \begin{subfigure}[b]{0.15\textwidth}
         \centering
         \includegraphics[width=\textwidth]{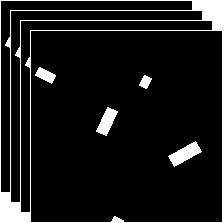}
         \caption{Vehicles}
         \label{fig:bev_vehicle}
     \end{subfigure}
     \hfil
     \begin{subfigure}[b]{0.15\textwidth}
         \centering
         \includegraphics[width=\textwidth]{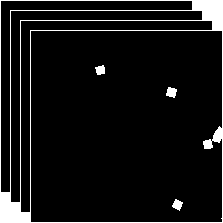}
         \caption{Pedestrians}
         \label{fig:bev_walker}
     \end{subfigure}
     \hfil
     \begin{subfigure}[b]{0.15\textwidth}
         \centering
         \includegraphics[width=\textwidth]{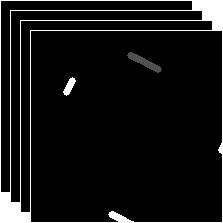}
         \caption{Lights and stops}
         \label{fig:bev_tl}
     \end{subfigure}
    \vspace{-1ex}
    \caption{\textbf{The BEV representation used by our Roach.}}
    \vspace{-2ex}
    \label{fig:bev}
\end{figure*}

\section{Method}
In this section we describe Roach and how IL agents can benefit from diverse supervisions supplied by Roach.

\subsection{RL Coach}
Our Roach has three features. 
Firstly, in contrast to previous RL agents, Roach does not depend on data from other experts.
Secondly, unlike the rule-based Autopilot, Roach is end-to-end trainable, hence it can generalize to new scenarios with minor engineering efforts. 
Thirdly, it has a high sample efficiency. 
Using our proposed input/output representation and exploration loss, training Roach from scratch to achieve top expert performance on the six LeaderBoard maps takes less than a week on a single GPU machine.

Roach consists of a policy network $\pi_\theta(\mathbf{a}| \mathbf{i}_\text{RL},\mathbf{m}_\text{RL})$ parameterized by $\theta$ and a value network $V_\phi(\mathbf{i}_\text{RL},\mathbf{m}_\text{RL})$ parameterized by $\phi$.
The policy network maps a BEV image $\mathbf{i}_\text{RL}$ and a measurement vector $\mathbf{m}_\text{RL}$ to a distribution of actions $\mathbf{a}$.
Finally the value network estimates a scalar value $v$, while taking the same inputs as the policy network.

\vspace{1ex}\noindent{\bf Input Representation:}
We use a BEV semantic segmentation image $\mathbf{i}_\text{RL} \in [0,1]^{W \times H \times C}$ to reduce the problem complexity, similar to the one used in \cite{DBLP:conf/rss/BansalKO19,chen2020learning,chen2019model}.
It is rendered using ground-truth simulation states and consists of $C$ grayscale images of size $W \times H$. 
The ego-vehicle is heading upwards and is centered in all images at $D$ pixels above the bottom, but it is not rendered.
Fig.~\ref{fig:bev} illustrates each channel of $\mathbf{i}_\text{RL}$. 
Drivable areas and intended routes are rendered respectively in Fig. \ref{fig:bev_road} and \ref{fig:bev_route}. 
In Fig. \ref{fig:bev_lane} solid lines are white and broken lines are grey. 
Fig. \ref{fig:bev_vehicle} is a temporal sequence of $K$ grayscale images in which cyclists and vehicles are rendered as white bounding boxes. 
Fig. \ref{fig:bev_walker} is the same as Fig. \ref{fig:bev_vehicle} but for pedestrians. 
Similarly, stop lines at traffic lights and trigger areas of stop signs are rendered in Fig. \ref{fig:bev_tl}. 
Red lights and stop signs are colored by the brightest level, yellow lights by an intermediate level and green lights by a darker level. 
A stop sign is rendered if it is active, i.e. the ego-vehicle enters its vicinity and disappears once the ego-vehicle has made a full stop.
By letting the BEV representation memorize if the ego-vehicle has stopped, we can use a network architecture without recurrent structure and hence reduce the model size of Roach.
A colored combination of all channels is visualized in Fig. \ref{fig:intro}.
We also feed Roach a measurement vector $\mathbf{m}_\text{RL} \in \mathbb{R}^6$ containing the states of the ego-vehicle not represented in the BEV, these include ground-truth measurements of steering, throttle, brake, gear, lateral and horizontal speed.

\vspace{1ex}\noindent{\bf Output Representation:}
Low-level actions of CARLA are $steering \in [-1,1]$, $throttle \in [0,1]$ and $brake \in [0,1]$. 
An effective way to reduce the problem complexity is predicting waypoint plans which are then tracked by a PID-controller to produce low-level actions \cite{chen2020learning,Rhinehart2020Deep}. 
However, a PID-controller is not reliable for trajectory tracking and requires excessive parameter tuning. 
A model-based controller would be a better solution, but CARLA's vehicle dynamics model is not directly accessible. 
To avoid parameter tuning and system identification, Roach directly predicts action distributions. 
Its action space is $\mathbf{a}\in [-1,1]^2$ for steering and acceleration, where positive acceleration corresponds to throttle and negative corresponds to brake.
To describe actions we use the Beta distribution $\mathcal{B}(\alpha,\beta)$, 
where $\alpha,\beta>0$ are respectively the concentration on $1$ and $0$.
Compared to the Gaussian distribution, which is commonly used in model-free RL, the support of the Beta distribution is bounded, thus avoiding clipping or squashing to enforce input constraints. This results in a better behaved learning problem since no $\text{tanh}$ layers are needed and the entropy and KL-divergence can be computed explicitly. 
Further, the modality of the Beta distribution is also suited for driving, where extreme maneuvers may often be taken, for example, emergency braking or taking a sharp turn.

\vspace{1ex}\noindent{\bf Training:}
We use proximal policy optimization (PPO) \cite{schulman2017proximal} with clipping to train the policy network $\pi_\theta$ and the value network $V_\phi$.
To update both networks, we collect trajectories by executing $\pi_{\theta_k}$ on CARLA. 
A trajectory $\tau =\{({\mathbf{i}_{\text{RL},k}},{\mathbf{m}_{\text{RL},k}},\mathbf{a}_k,r_k)^{T}_{k=0}, z \}$ includes BEV images $\mathbf{i}_\text{RL}$, measurement vectors $\mathbf{m}_\text{RL}$, actions $\mathbf{a}$, rewards $r$ and a terminal event $z \in \mathcal{Z}$ that triggers the termination of an episode. 
The value network is trained to regress the expected returns, whereas the policy network is updated via
\begin{equation}
    \theta_{k+1} = \arg \max_\theta \mathop{\text{E}}_{\tau \sim \pi_{\theta_k}} \left[\mathcal{L}_\text{ppo}+ \mathcal{L}_\text{ent}+ \mathcal{L}_\text{exp}\right].
\end{equation}
The first objective $\mathcal{L}_\text{ppo}$ is the clipped policy gradient loss with advantages estimated using generalized advantage estimation \cite{schulman2016gae}.
The second objective $\mathcal{L}_\text{ent}$ is a maximum entropy loss commonly employed to encourage exploration
\begin{equation}
    \mathcal{L}_\text{ent} = -\lambda_\text{ent} \cdot \text{H} \left(
    \pi_\theta(\cdot | \mathbf{i}_\text{RL},\mathbf{m}_\text{RL}) 
    \right).
\end{equation}
Intuitively $\mathcal{L}_\text{ent}$ pushes the action distribution towards a uniform prior because maximizing entropy is equivalent to minimizing the KL-divergence to a uniform distribution,
\begin{equation}
    \text{H} \left( \pi_\theta  \right)= -
    \text{KL}\left( 
    \pi_\theta \parallel \mathcal{U}(-1,1) 
    \right),
\end{equation}
if both distributions share the same support.
This inspires us to propose a generalized form of $\mathcal{L}_\text{ent}$, which encourages exploration in sensible directions that comply with basic traffic rules. We call it the exploration loss and define it as
\begin{equation}
\begin{split}
    \mathcal{L}_\text{exp}  = \, & \lambda_\text{exp} \cdot
    \mathbbm{1}_{ \{T-N_{z}+1,\dots,T \} }(k) \\
     \cdot
    & \text{KL}( 
    \pi_\theta(\cdot|{\mathbf{i}_{\text{RL},k}},{\mathbf{m}_{\text{RL},k}} )\parallel p_{z}),
\end{split}
\end{equation}
where $\mathbbm{1}$ is the indicator function and $z \in \mathcal{Z}$ is the event that ends the episode. The terminal condition set $\mathcal{Z}$ includes collision, running traffic light/sign, route deviation and being blocked. 
Unlike $\mathcal{L}_\text{ent}$ which imposes a uniform prior on the actions at all time steps regardless of which $z$ is triggered, $\mathcal{L}_\text{exp}$
shifts actions within the last $N_{z}$ steps of an episode towards a predefined exploration prior $p_z$ which encodes an ``advice" to prevent the triggered event $z$ from happening again.
In practice, we use $N_z=100,\forall z \in \mathcal{Z}$.
If $z$ is related to a collision or running traffic light/sign, we apply $p_z=\mathcal{B}(1,2.5)$ on the acceleration to encourage Roach to slow down while the steering is unaffected. 
In contrast, if the car is blocked we use an acceleration prior $\mathcal{B}(2.5,1)$. 
For route deviations, a uniform prior $\mathcal{B}(1,1)$ is applied on the steering.
Despite being equivalent to maximizing entropy in this case, the exploration loss further encourages exploration on steering angles during the last 10 seconds before the route deviation. 




\begin{figure*}[t]
     \centering
     \begin{subfigure}[b]{0.49\textwidth}
         \centering
         \includegraphics[width=\textwidth]{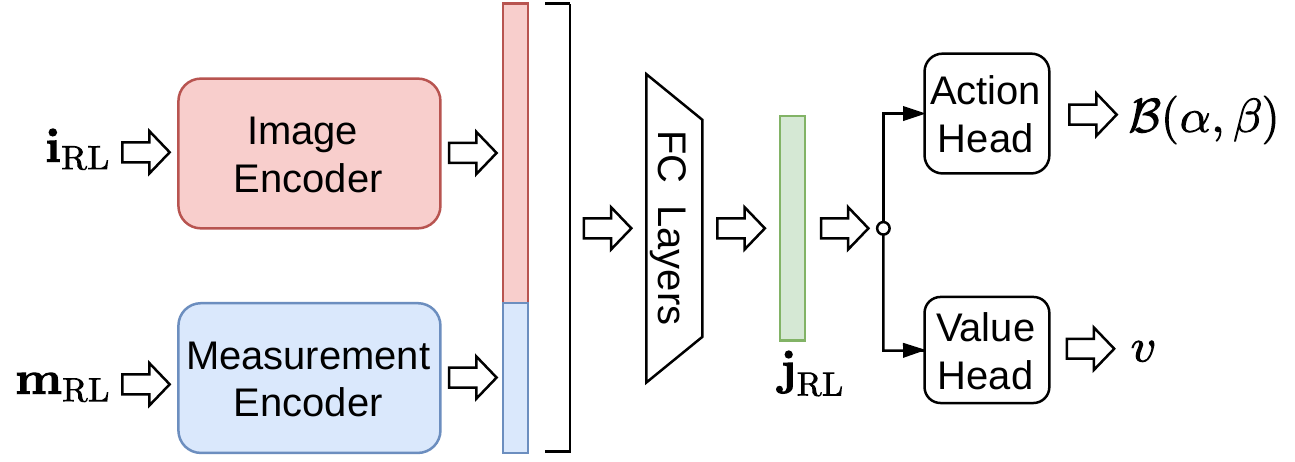}
         \caption{Roach}
         \label{fig:net_ppo}
     \end{subfigure}
     \hfill
     \begin{subfigure}[b]{0.47\textwidth}
         \centering
         \includegraphics[width=\textwidth]{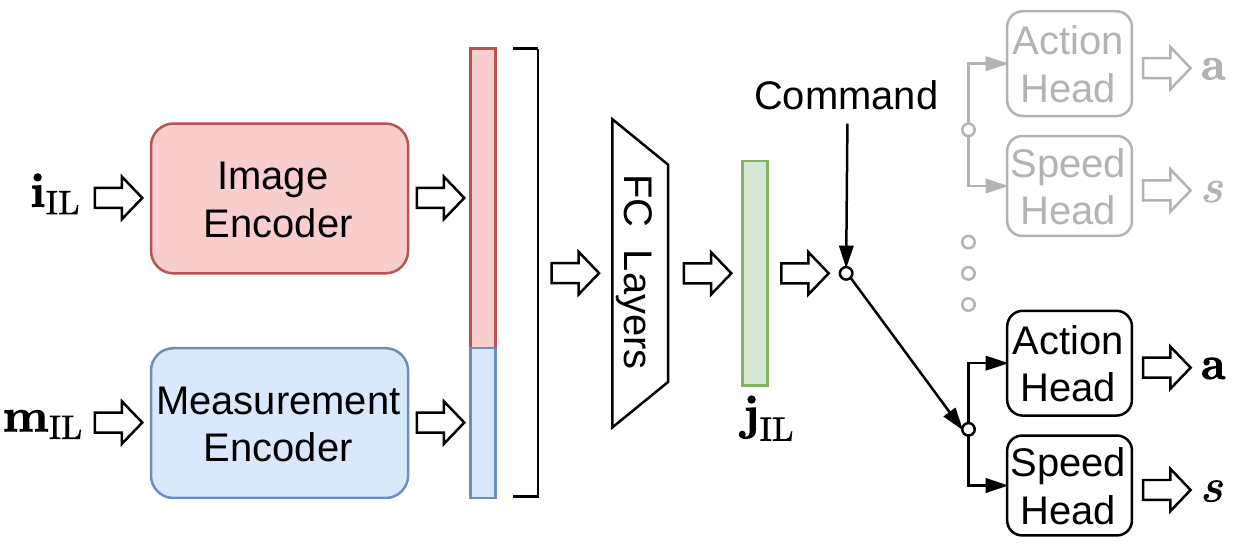}
         \caption{CILRS}
         \label{fig:net_cilrs}
     \end{subfigure}
    \vspace{-1ex}
    \caption{\textbf{Network architecture of Roach, the RL expert, and CILRS, the IL agent.}}
    \vspace{-2ex}
    \label{fig:network}
\end{figure*}

\vspace{1ex}\noindent{\bf Implementation Details:}
Our implementation of PPO-clip is based on \cite{stable-baselines3} and the network architecture is illustrated in Fig.~\ref{fig:net_ppo}. 
We use six convolutional layers to encode the BEV and two fully-connected (FC) layers to encode the measurement vector. 
Outputs of both encoders are concatenated and then processed by another two FC layers to produce a latent feature $\mathbf{j}_\text{RL}$, which is then fed into a value head and a policy head, each with two FC hidden layers. 
Trajectories are collected from six CARLA servers at 10 FPS, each server corresponds to one of the six LeaderBoard maps. 
At the beginning of each episode, a pair of start and target location is randomly selected and the desired route is computed using the A$^*$ algorithm.
Once the target is reached, a new random target will be chosen, hence the episode is endless unless one of the terminal conditions in $\mathcal{Z}$ is met. 
We use the reward of \cite{toromanoff2019deep} and additionally penalize large steering changes to prevent oscillating maneuvers. 
To avoid infractions at high speed, we add an extra penalty proportional to the ego-vehicle's speed.
More details are in the supplement.

\subsection{IL Agents Supervised by Roach}
To allow IL agents to benefit from the informative supervisions generated by Roach, we formulate a loss for each of the supervisions.
Our training scheme using Roach can be applied to improve the performance of existing IL agents.
Here we use DA-RB \cite{prakash2020exploring} (CILRS \cite{codevilla2019exploring} + DAGGER \cite{ross2011reduction}) as an example to demonstrate its effectiveness.


\vspace{1ex}\noindent{\bf CILRS:} 
The network architecture of CILRS is illustrated in Fig. \ref{fig:net_cilrs}, it includes a perception module that encodes the camera image $\mathbf{i}_\text{IL}$ and a measurement module that encodes the measurement vector $\mathbf{m}_\text{IL}$. 
Outputs of both modules are concatenated and processed by FC layers to generate a bottleneck latent feature $\mathbf{j}_\text{IL}$. 
Navigation instructions are given as discrete high-level commands and for each kind of command a branch is constructed. 
All branches share the same architecture, while each branch contains an action head predicting continuous actions $\mathbf{a}$ and a speed head predicting the current speed $s$ of the ego-vehicle. 
The latent feature $\mathbf{j}_\text{IL}$ is processed by the branch selected by the command. 
The imitation objective of CILRS consists of an L1 action loss
\begin{equation} \label{eq: loss_a}
    \mathcal{L}_\text{A} = \|\hat{\mathbf{a}} - \mathbf{a} \|_1
\end{equation}
and a speed prediction regularization
\begin{equation} \label{loss:s}
    \mathcal{L}_\text{S} = \lambda_\text{S} \cdot |\hat{s}-s |, 
\end{equation}
where $\lambda_\text{s}$ is a scalar weight, $\hat{\mathbf{a}}$ is the expert's action, $\hat{s}$ is the measured speed, $\mathbf{a}$ and $s$ are action and speed predicted by CILRS. 
Expert actions $\hat{\mathbf{a}}$ may come from the Autopilot, which directly outputs deterministic actions, or from Roach, where the distribution mode is taken as the deterministic output. Besides deterministic actions, Roach also predicts action distributions, values and latent features. 
Next we will formulate a loss function for each of them.

\vspace{1ex}\noindent{\bf Action Distribution Loss:} 
Inspired by \cite{hinton2015distilling} which suggests soft targets may provide more information per sample than hard targets, we propose a new action loss based on the action distributions as a replacement for $\mathcal{L}_\text{A}$.
The action head of CILRS is modified to predict distribution parameters and the loss is formulated as a KL-divergence
\begin{equation} \label{eq: loss_kl}
    \mathcal{L}_\text{K} = \text{KL}(\hat{\pi} \| \pi)
\end{equation}
between the action distribution $\hat{\pi}$ predicted by the Roach expert and $\pi$ predicted by the CILRS agent.

\vspace{1ex}\noindent{\bf Feature Loss:} 
Feature matching is an effective way to transfer knowledge between networks and its effectiveness in supervising IL driving agents is demonstrated in \cite{hou2019learning,zhao2020sam}.
The latent feature $\mathbf{j}_\text{RL}$ of Roach is a compact representation that contains essential information for driving as it can be mapped to expert actions using an action head consists of only two FC layers (cf. Fig.~\ref{fig:net_ppo}).
Moreover, $\mathbf{j}_\text{RL}$ is invariant to rendering and weather as Roach uses the BEV representation.
Learning to embed camera images to the latent space of $\mathbf{j}_\text{RL}$ should help IL agents to generalize to new weather and new situations.
Hence, we propose the feature loss
\begin{equation} \label{eq: loss_f}
    \mathcal{L}_\text{F} = \lambda_\text{F} \cdot \| \mathbf{j}_\text{RL} - \mathbf{j}_\text{IL} \|_2^2.
\end{equation}
\noindent{\bf Value Loss:} 
Multi-task learning with driving-related side tasks could also boost the performance of end-to-end IL driving agents as shown in \cite{xu2017end}, which used scene segmentation as a side task.
Intuitively, the value predicted by Roach contains driving relevant information because it estimates the expected future return, which relates to how dangerous a situation is. 
Therefore, we augment CILRS with a value head and regress value as a side task. 
The value loss is the mean squared error between $\hat{v}$, the value estimated by Roach, and $v$, the value predicted by CILRS,
\begin{equation} \label{eq: loss_v}
    \mathcal{L}_\text{V} = \lambda_\text{V} \cdot (\hat{v}-v)^2.
\end{equation}
\noindent{\bf Implementation Details:}
Our implementation follows DA-RB \cite{prakash2020exploring}. 
We choose a Resnet-34 pretrained on ImageNet as the image encoder to generate a 1000-dimensional feature given $\mathbf{i}_\text{RL} \in [0,1]^{900 \times 256 \times 3}$, a wide-angle camera image with a $100^{\circ}$ horizontal FOV. 
Outputs of the image and the measurement encoder are concatenated and processed by three FC layers to generate $\mathbf{j}_\text{IL} \in \mathbb{R}^{256}$, which shares the same size as $\mathbf{j}_\text{RL}$.
More details are found in the supplement.

\section{Experiments}
\noindent{\bf Benchmarks:} 
All evaluations are completed on CARLA 0.9.11.
We evaluate our methods on the NoCrash \cite{codevilla2019exploring} and the offline LeaderBoard benchmark\footnote{In contrast to the Leaderboard online ranking, this benchmark is evaluated offline on the Leaderboard public routes (50 training, 26 testing).} \cite{leaderboard}.
Each benchmark specifies its training towns and weather, where the agent is allowed to collect data, and evaluates the agent in new towns and weather.
The NoCrash benchmark considers generalization from Town 1, a European town composed of solely one-lane roads and T-junctions, to Town 2, a smaller version of Town 1 with different textures. 
By contrast, the LeaderBoard considers a more difficult generalization task in six maps that cover diverse traffic situations, including freeways, US-style junctions, roundabouts, stop signs, lane changing and merging.
Following the NoCrash benchmark, we test generalization from four training weather types to two new weather types.
But to save computational resources, only two out of the four training weather types are evaluated.
The NoCrash benchmark comes with three levels of traffic density (empty, regular and dense), which defines the number of pedestrians and vehicles in each map.
We focus on the NoCrash-dense and introduce a new level between regular and dense traffic, NoCrash-busy, to avoid congestion that often appears in the dense traffic setting.
For the offline LeaderBoard the traffic density in each map is tuned to be comparable to the busy traffic setting. 

\vspace{1ex}\noindent{\bf Metrics:} 
Our results are reported in success rate, the metric proposed by NoCrash, and driving score, a new metric introduced by the CARLA LeaderBoard. 
The success rate is the percentage of routes completed without collision or blockage. 
The driving score is defined as the product of route completion, the percentage of route distance completed, and infraction penalty, a discount factor that aggregates all triggered infractions.
For example, if the agent ran two red lights in one route and the penalty coefficient for running one red light was $0.7$, then the infraction penalty would be  $0.7^{2}$$=$$0.49$.
Compared to the success rate, the driving score is a fine-grained metric that considers more kinds of infractions and it is better suited to evaluate long-distance routes.
More details about the benchmarks and the complete results are found in the supplement.


\begin{figure}[t]
\begin{center}
    \includegraphics[width=\linewidth]{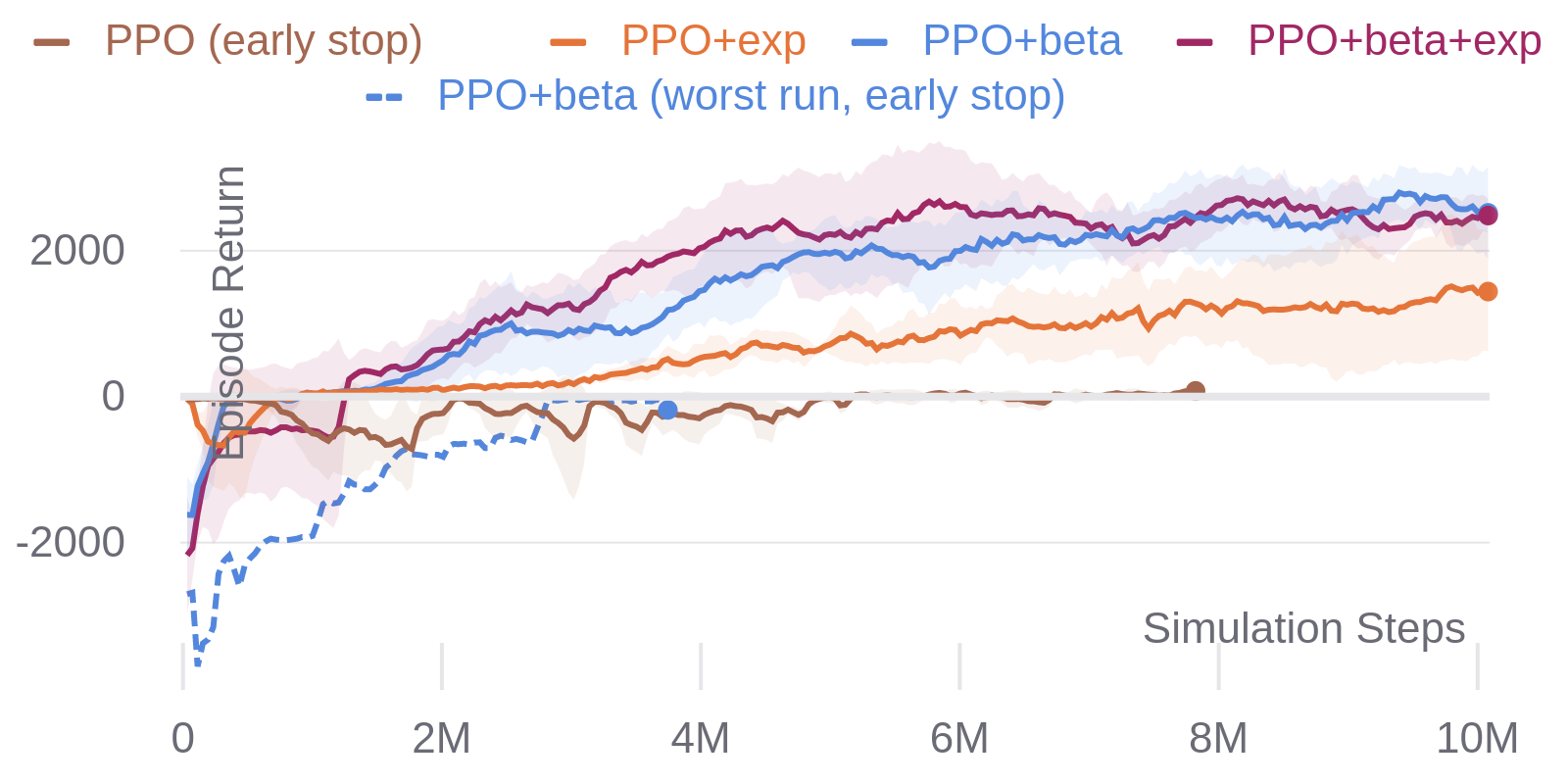}
\end{center}
\vspace{-4ex}
\caption{\textbf{Learning curves of RL experts }
trained in CARLA Town 1-6.
Solid lines show the mean and shaded areas show the standard deviation of episode returns across 3 seeds. The dashed line shows an outlier run that collapsed.}
\vspace{-2ex}
\label{fig:rl}
\end{figure}

\subsection{Performance of Experts}
We use CARLA 0.9.10.1 to train RL experts and fine-tune our Autopilot, yet all evaluations are still on 0.9.11.

\vspace{1ex}\noindent{\bf Sample Efficiency:}
To improve the sample efficiency of PPO, we propose to use BEV instead of camera images, Beta instead of Gaussian distributions, and the exploration loss in addition to the entropy loss.
Since the benefit of using a BEV representation is obvious, here we only ablate the Beta distribution and the exploration loss.
As shown in Fig.~\ref{fig:rl}, the baseline PPO with Gaussian distribution and entropy loss is trapped in a local minimum where staying still is the most rewarding strategy.
Leveraging the exploration loss, PPO+exp can be successfully trained despite relatively high variance and low sample efficiency.
The Beta distribution helps substantially, but without the exploration loss the training still collapsed in some cases due to insufficient exploration (cf. dashed blue line in Fig.~\ref{fig:rl}).
Our Roach (PPO+beta+exp) uses both Beta distribution and exploration loss to ensure stable and sample efficient training.
The training takes around 1.7M steps in each of the six CARLA servers, this accounts for 10M steps in total, which takes roughly a week on an AWS EC2 g4dn.4xlarge or 4 days on a 2080 Ti machine with 12 cores.

\vspace{1ex}\noindent{\bf Driving Performance:}
Table~\ref{table:expert_performance} compares different experts on the NoCrash-dense and on all 76 LeaderBoard routes under dynamic weather with busy traffic.
Our Autopilot is a strong baseline expert that achieves a higher success rate than the Autopilot used in LBC and DA-RB.
We evaluate three RL experts - 
(1) Roach, the proposed RL coach using Beta distribution and exploration prior.
(2) PPO+beta, the RL coach trained without using the exploration prior. 
(3) PPO+exp, the RL coach trained without using the Beta distribution.
In general, our RL experts achieve comparable success rates and higher driving scores than Autopilots because RL experts handle traffic lights in a better way (cf. Table~\ref{table:infraction}).
The two Autopilots often run red lights because they drive over-conservatively and wait too long at the junction, thus missing the green light.
Among RL experts, PPO+beta and Roach, the two RL experts using a Beta distribution, achieve the best performance, while the difference between both is not significant. PPO+exp performs slightly worse, but it still achieves better driving scores than our Autopilot. 

\begin{table}
\setlength{\tabcolsep}{2.32pt}
\centering
\begin{tabular}{lccccc}
\toprule
Suc. Rate \% $\uparrow$
& NCd-tt & NCd-tn  & NCd-nt & NCd-nn & LB-all  \\ 
\cmidrule(lr){1-1}\cmidrule(lr){2-6}
PPO+exp & $86 \pm 6$ & $86 \pm 6$ & $79 \pm 6$ & $77 \pm 5$ & $67\pm3$  \\
PPO+beta & $\mathbf{95} \pm 3$ & $\mathbf{95} \pm 3$ & $83 \pm 5$ & $\mathbf{87} \pm 6$ & $72 \pm 5$  \\
Roach & $91 \pm 4$ & $90 \pm 7$ & $\mathbf{83} \pm 3$ & $83 \pm 3$ & $72 \pm 6$  \\
\cmidrule(lr){1-1}\cmidrule(lr){2-6}
AP (ours) & 
$\mathbf{95} \pm 3$ & $\mathbf{95} \pm 3$ & $83 \pm 5$ & $81 \pm 2$ & $\mathbf{75} \pm 8$ \\
AP-lbc \cite{chen2020learning}
& $86 \pm 3$ & $83 \pm 6$ & $60 \pm 3$ & $59 \pm 8$ & N/A \\
AP-darb \cite{prakash2020exploring}
& $71 \pm 4$ & $72 \pm 3$ & $41 \pm 2$ & $43 \pm 2$ & N/A \\
\toprule
Dri. Score \% $\uparrow$
& NCd-tt & NCd-tn  & NCd-nt & NCd-nn & LB-all  \\ 
\cmidrule(lr){1-1}\cmidrule(lr){2-6}
PPO+exp & $92 \pm 2$ & $92 \pm 2$ & $88 \pm 3$ & $86 \pm 1$ & $ 83\pm0$  \\
PPO+beta & $\mathbf{98} \pm 2$ & $\mathbf{98} \pm 2$ & $90 \pm 3$ & $\mathbf{92} \pm 2$ & $\mathbf{86} \pm 2$  \\
Roach & $95 \pm 2$ & $95 \pm 3$ & $\mathbf{91} \pm 3$ & $90 \pm 2$ & $85 \pm 3$  \\
\cmidrule(lr){1-1}\cmidrule(lr){2-6}
AP (ours)
& $86 \pm 2$ & $86 \pm 2$ & $70 \pm 2$ & $70 \pm 1$
& $78 \pm 3$ \\
\bottomrule
\end{tabular}
\vspace{-1ex}
\caption{\textbf{Success rate and driving score of experts.} Mean and standard deviation over 3 evaluation seeds. NCd: NoCrash-dense. tt: train town \& weather. tn: train town \& new weather. nt: new town \& train weather. nn: new town \& weather. LB-all: all 76 routes of LeaderBoard with dynamic weather. AP: CARLA Autopilot. For RL experts the best checkpoint among all training seeds and runs is used.}
\label{table:expert_performance}
\vspace{-2ex}
\end{table}

\subsection{Performance of IL Agents}
\begin{figure*}[t]
     \centering
     \includegraphics[width=0.99\textwidth]{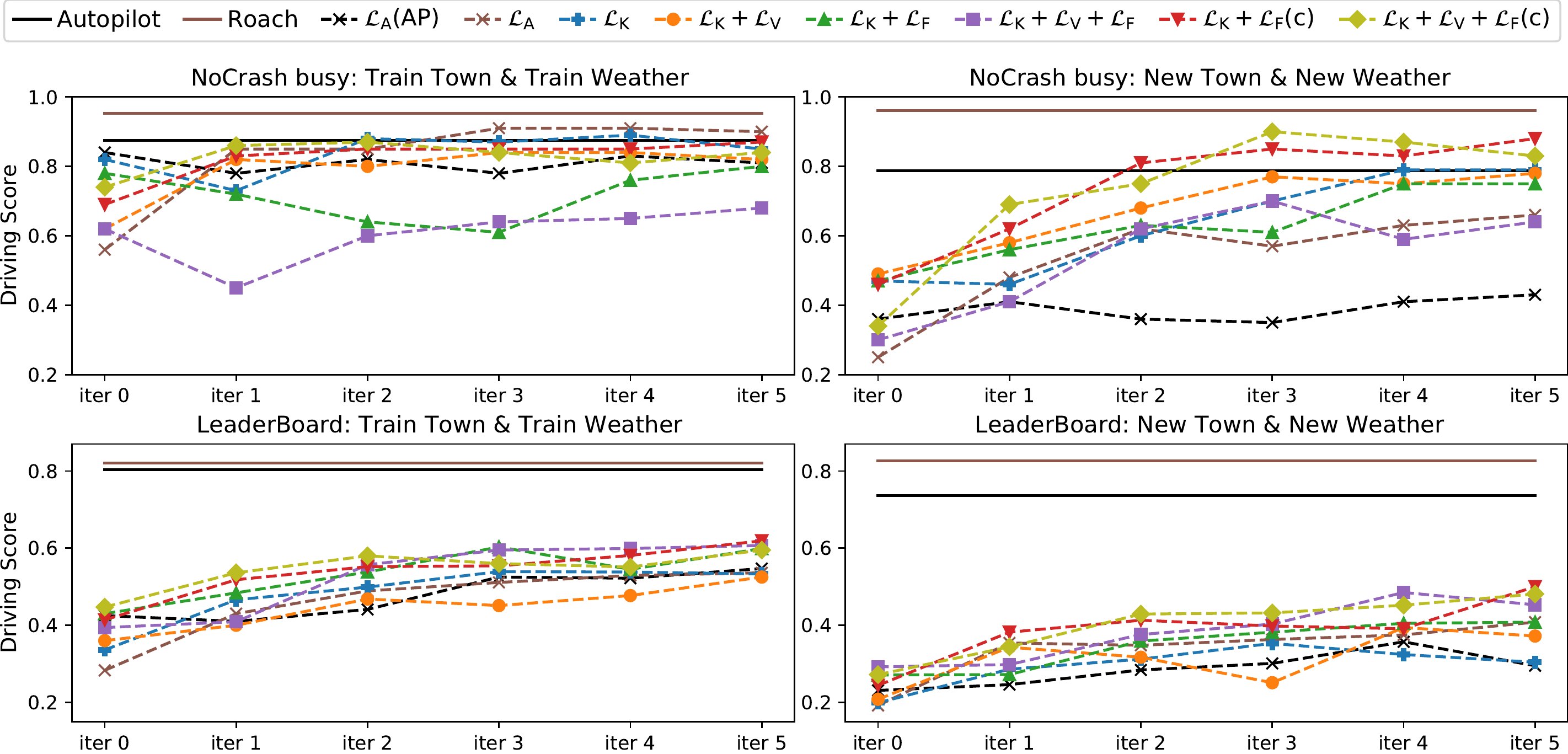}
     \vspace{-1ex}
     \caption{\textbf{Driving score of experts and IL agents.} All IL agents (dashed lines) are supervised by Roach except for $\mathcal{L}_\text{A}(\text{AP})$, which is supervised by our Autopilot. For IL agents at the 5th iteration on NoCrash and all experts, results are reported as the mean over 3 evaluation seeds. Others are evaluated with one seed. The offline Leaderboard benchmark is used here.}
     \vspace{-1.5ex}
    \label{fig:score_eu_lb_tt_tn}
\end{figure*}

The performance of an IL agent is limited by the performance of the expert it is imitating.
If the expert performs poorly, it is not sensible to compare IL agents imitating that expert.
As shown in Table~\ref{table:expert_performance}, this issue is evident in the NoCrash new town with dense traffic, where Autopilots do not perform well. 
To ensure a high performance upper-bound and hence a fair comparison, we conduct ablation studies (Fig.~\ref{fig:score_eu_lb_tt_tn} and Table~\ref{table:infraction}) under the busy traffic setting such that our Autopilot can achieve a driving score of 80\% and a success rate of 90\%. In order to compare with the state-of-the-art, the best model from the ablation studies is still evaluated on NoCrash with dense traffic in Table~\ref{table:sucess_rate_nc_dense}.

The input measurement vector $\mathbf{m}_\text{IL}$ is different for the NoCrash and for the LeaderBoard. For NoCrash, $\mathbf{m}_\text{IL}$ is just the speed.
For the LeaderBoard, $\mathbf{m}_\text{IL}$ contains additionally a 2D vector pointing to the next desired waypoint.
This vector is computed from noisy GPS measurements and the desired route is specified as sparse GPS locations.
The LeaderBoard instruction suggests that it is used to disambiguate situations where the semantics of left and right are not clear due to the complexity of the considered map.

\begin{table}
\setlength{\tabcolsep}{2.67pt}
\centering
\begin{tabular}{lccccc}
\toprule
Success Rate \% $\uparrow$
&  NCd-tt & NCd-tn  & NCd-nt & NCd-nn  \\ 
\cmidrule(lr){1-1}\cmidrule(lr){2-5}
LBC \cite{chen2020learning} (0.9.6) & 
$71 \pm 5$ & $63 \pm 3$ & $51 \pm 3$ & $39 \pm 6$ \\
SAM \cite{zhao2020sam} (0.8.4) & 
$54 \pm 3$ & $47 \pm 5$ & $29 \pm 3$ & $29 \pm 2$ \\
LSD \cite{ohn2020learning} (0.8.4) & 
N/A & N/A & $30 \pm 4$ & $32 \pm 3$ \\
DA-RB\textsuperscript{+}(E) \cite{prakash2020exploring} & 
$66 \pm 5$ & $56 \pm 1$ & $36 \pm 3$ & $35 \pm 2$ \\
DA-RB\textsuperscript{+} \cite{prakash2020exploring} (0.8.4)  & 
$62 \pm 1$ & $60 \pm 1$ & $34 \pm 2$ & $25 \pm 1$ \\
Our baseline, $\mathcal{L}_\text{A}\text{(AP)}$ & 
$\mathbf{88} \pm 4$ & $29 \pm 3$ & $32 \pm 11$ & $28 \pm 4$ \\
Our best, $\mathcal{L}_\text{K}+\mathcal{L}_\text{F}(\text{c})$ & 
$86 \pm 5$ & $\mathbf{82} \pm 2$ & $\mathbf{78} \pm 5$ & $\mathbf{78} \pm 0$ \\
\bottomrule
\end{tabular}
\vspace{-1ex}
\caption{\textbf{Success rate of camera-based end-to-end IL agents on NoCrash-dense.} Mean and standard deviation over 3 seeds. Our models are from DAGGER iteration 5. For DA-RB, + means triangular perturbations are added to the off-policy dataset, (E) means ensemble of all iterations.}
\label{table:sucess_rate_nc_dense}
\vspace{-2ex}
\end{table}
\begin{table*}
\setlength{\tabcolsep}{3.8pt}
\centering
\begin{tabular}{lccccccccc} 
\toprule
& \begin{tabular}{@{}c@{}}Success \\ rate \end{tabular} 
& \begin{tabular}{@{}c@{}}Driving \\ score \end{tabular} 
& \begin{tabular}{@{}c@{}}Route \\ compl. \end{tabular} 
& \begin{tabular}{@{}c@{}}Infrac. \\ penalty \end{tabular} 
& \begin{tabular}{@{}c@{}}Collision \\ others \end{tabular} 
& \begin{tabular}{@{}c@{}}Collision \\ pedestrian \end{tabular} 
& \begin{tabular}{@{}c@{}}Collision \\ vehicle \end{tabular}  
& \begin{tabular}{@{}c@{}}Red light \\ infraction \end{tabular}  
& \begin{tabular}{@{}c@{}}Agent \\ blocked \end{tabular}  \\
\cmidrule(lr){1-1}\cmidrule(lr){2-5}\cmidrule(lr){6-10}
iter 5
& \%, $\uparrow$
& \%, $\uparrow$
& \%, $\uparrow$
& \%, $\uparrow$
& \#/Km, $\downarrow$
& \#/Km, $\downarrow$
& \#/Km, $\downarrow$
& \#/Km, $\downarrow$
& \#/Km, $\downarrow$
\\
\cmidrule(lr){1-1}\cmidrule(lr){2-5}\cmidrule(lr){6-10}
$\mathcal{L}_\mathrm{A}(\text{AP})$
& $31 \pm 7$ & $43 \pm 2$ & $62 \pm 6$ & $77 \pm 4$ 
& $0.54 \pm 0.53$ & $\mathbf{0}\pm0$ & $0.63 \pm 0.50$ & $3.33 \pm 0.58$ & $19.4\pm 14.4$ \\
$\mathcal{L}_\text{A}$
& $57\pm7$ & $66\pm3$ & $84\pm3$ & $76\pm1$ 
& $2.07\pm1.37$ & $\mathbf{0}\pm0$ & $1.36\pm1.10$ & $1.4\pm0.2$ & $2.82\pm1.45$ \\
$\mathcal{L}_\text{K}$
& $74\pm3$ & $79\pm0$ & $91\pm2$ & $86\pm1$ 
& $0.50\pm0.25$ & $\mathbf{0}\pm0$ & $0.53\pm0.18$ & $0.68\pm0.08$ & $3.39\pm0.20$ \\
$\mathcal{L}_\text{K}+\mathcal{L}_\text{F}(\text{c})$
& $\mathbf{87} \pm 5$ & $\mathbf{88} \pm 3$ & $\mathbf{96} \pm 0$ & $\mathbf{91} \pm 3$ 
& $\mathbf{0.08} \pm 0.04$ & $0.01 \pm 0.02$ & $\mathbf{0.23} \pm 0.08$ & $\mathbf{0.61} \pm 0.23$ & $\mathbf{0.84} \pm 0.04$ \\
\cmidrule(lr){1-1}\cmidrule(lr){2-5}\cmidrule(lr){6-10}
Roach
& $95 \pm 2$ & $96 \pm 3$ & $100 \pm 0$ & $96 \pm 3$ 
& $0 \pm 0$ & $0.11 \pm 0.07$ & $0.04 \pm 0.05$ & $0.16 \pm 0.20$ & $0 \pm 0$ \\
Autopilot
& $91 \pm 1$ & $79 \pm 2$ & $98 \pm 1$ & $80 \pm 2$ 
& $0 \pm 0$ & $0 \pm 0$ & $0.18 \pm 0.08$ & $1.93 \pm 0.23$ & $0.18 \pm 0.08$\\
\bottomrule
\end{tabular}
\vspace{-1ex}
\caption{\textbf{Driving performance and infraction analysis of IL agents on NoCrash-busy, new town \& new weather.} Mean and standard deviation over 3 evaluation seeds.}
\vspace{-2.5ex}
\label{table:infraction}
\end{table*}

\vspace{1ex}\noindent{\bf Ablation:}
Fig.~\ref{fig:score_eu_lb_tt_tn} shows driving scores of experts and IL agents at each DAGGER iteration on NoCrash and offline LeaderBoard with busy traffic.
The baseline $\mathcal{L}_\text{A}(\text{AP})$ is our implementation of DA-RB\textsuperscript{+} supervised by our Autopilot. 
Given our improved Autopilot, it is expected that $\mathcal{L}_\text{A}(\text{AP})$ can achieve higher success rates than those reported in the DA-RB paper, but this is not observed in Table~\ref{table:sucess_rate_nc_dense}.
The large performance gap between the Autopilot and $\mathcal{L}_\text{A}(\text{AP})$ (cf. Fig.~\ref{fig:score_eu_lb_tt_tn}), especially while generalizing to a new town and new weather, indicates the limitation of this baseline.

By replacing the Autopilot with Roach, $\mathcal{L}_\text{A}$ performs better overall than $\mathcal{L}_\text{A}(\text{AP})$.
Further learning from the action distribution, $\mathcal{L}_\text{K}$ generalizes better than $\mathcal{L}_\text{A}$ on the NoCrash but not on the offline LeaderBoard.
Feature matching only helps when $\mathbf{j}_\text{IL}$ is provided with the necessary information needed to reproduce $\mathbf{j}_\text{RL}$.
In our case, $\mathbf{j}_\text{RL}$ contains navigational information as the desired route is rendered in the BEV input.
For the LeaderBoard, navigational information is partially encoded in $\mathbf{m}_\text{IL}$, which includes the vector to the next desired waypoint, so better performance is observed by using $\mathcal{L}_\text{F}$.
But for NoCrash this information is missing as $\mathbf{m}_\text{IL}$ is just the speed, hence it is impractical for $\mathbf{j}_\text{IL}$ to mimic $\mathbf{j}_\text{RL}$ and this causes the inferior performance of $\mathcal{L}_\text{K}+\mathcal{L}_\text{F}$ and $\mathcal{L}_\text{K}+\mathcal{L}_\text{F}+\mathcal{L}_\text{V}$.
To confirm this hypothesis, we evaluate a single-branch network architecture where the measurement vector $\mathbf{m}_\text{IL}$ is augmented by the command encoded as a one-hot vector.
Using feature matching with this architecture, $\mathcal{L}_\text{K}+\mathcal{L}_\text{F}(\text{c})$ and $\mathcal{L}_\text{K}+\mathcal{L}_\text{V}+\mathcal{L}_\text{F}(\text{c})$ achieve the best driving score among IL agents in the NoCrash new town \& weather generalization test, even outperforming the Autopilot.

Using value supervision in addition to feature matching helps the DAGGER process to converge faster as shown by $\mathcal{L}_\text{K}+\mathcal{L}_\text{V}+\mathcal{L}_\text{F}$ and $\mathcal{L}_\text{K}+\mathcal{L}_\text{V}+\mathcal{L}_\text{F}(\text{c})$.
However, without feature matching, using value supervision alone $\mathcal{L}_\text{K}+\mathcal{L}_\text{V}$ does not demonstrate superior performance.
This indicates a potential synergy between feature matching and value estimation.
Intuitively, the latent feature of Roach encodes the information needed for value estimation, hence mimicking this feature should help to predict the value,
while value estimation could help to regularize feature matching.


\vspace{1ex}\noindent{\bf Comparison with the State-of-the-art:}
In Table~\ref{table:sucess_rate_nc_dense} we compare the baseline $\mathcal{L}_\text{A}(\text{AP})$ and our best performing agent $\mathcal{L}_\text{K}+\mathcal{L}_\text{F}(\text{c})$ with the state-of-the-art on the NoCrash-dense benchmark.
Our $\mathcal{L}_\text{A}(\text{AP})$ performs comparably to DA-RB\textsuperscript{+} except when generalizing to the new weather, where there is an incorrect rendering of after-rain puddles on CARLA 0.9.11 (see supplement for visualizations).
This issue does not affect our best method $\mathcal{L}_\text{K}+\mathcal{L}_\text{F}(\text{c})$ due to the stronger supervision of Roach. 
By mimicking the weather-agnostic Roach, the performance of our IL agent drops by less than $10\%$ while generalizing to the new town and weather.
Hence if the Autopilot is considered the performance upper-bound, it is fair to claim our approach saturates the NoCrash benchmark.
However, as shown in Fig.~\ref{fig:score_eu_lb_tt_tn}, there is still space for improvement on NoCrash compared to Roach and the performance gap on the offline LeaderBoard highlights the importance of this new benchmark.

\vspace{1ex}\noindent{\bf Performance and Infraction Analysis:}
Table~\ref{table:infraction} provides the detailed performance and infraction analysis on the NoCrash benchmark with busy traffic in the new town \& weather setting.
Most notably, the extremely high ``Agent blocked'' of our baseline $\mathcal{L}_\text{A}(\text{AP})$ is due to reflections from after-rain puddles.
This problem is largely alleviated by imitating Roach, which drives more naturally, and $\mathcal{L}_\text{A}$ shows an absolute improvement of $23\%$ in terms of driving score.
In other words this is the gain achieved by using a better expert, but the same imitation learning approach. 
Further using the improved supervision from soft targets and latent features results in our best model $\mathcal{L}_\text{K}+\mathcal{L}_\text{F}(\text{c})$, which demonstrates another $22\%$ absolute improvement.
By handling red lights in a better way, this agent achieves $88\%$, an expert-level driving score, using a single camera image as input.

\section{Conclusion}
We present Roach, an RL expert, and an effective way to imitate this expert.
Using the BEV representation, Beta distribution and the exploration loss, Roach sets the new performance upper-bound on CARLA while demonstrating high sample efficiency.
To enable a more effective imitation, we propose to learn from soft targets, values and latent features generated by Roach.
Supervised by these informative targets, a baseline end-to-end IL agent using a single camera image as input can achieve state-of-the-art performance, even reaching expert-level performance on the NoCrash-dense benchmark.
Future works include performance improvement on simulation benchmarks and real-world deployment.
To saturate the LeaderBoard, the model capacity shall be increased \cite{DBLP:conf/rss/BansalKO19,hecker2018end,ohn2020learning}.
To apply Roach to label real-world on-policy data, several sim-to-real gaps have to be addressed besides the photorealism, which is partially alleviated by the BEV.
For urban driving simulators, the realistic behavior of road users is of utmost importance \cite{suo2021trafficsim}.
\noindent{\bf Acknowledgements:}
This work was funded by Toyota Motor Europe via the research project TRACE Zurich.


{\small
\bibliographystyle{ieee_fullname}
\bibliography{egbib}
}

\clearpage

\appendix

\section{Summary}
In the appendix, we provide (1) an overview of supplementary videos and codes, (2) implementation details of the RL experts and the IL agents, (3) details regarding benchmarks, and (4) additional experimental results.


\section{Other Supplementary Materials}

\subsection{Videos}

To investigate how different agents actually drive, we provide three videos.
\textbf{roach.mp4} shows the driving performance of Roach, and highlights that it has a natural driving style and that it can handle complex traffic scenes. 
In \textbf{autopilot.mp4} we demonstrate the rule-based CARLA Autopilot. This agent uses unnatural brake actuation, i.e. it only uses emergency braking. 
Further, this video also highlights that in dense traffic, the rule-based agent can get stuck due to conservative danger predictions.
For more details about the Autopilot and changes we made see Section \ref{sec:autopilot}. Finally, in \textbf{il\_agent.mp4} we demonstrate our best roach-supervised IL agent, showing that the agent can handle complex traffic scenes but also highlighting failure cases. 
In detail: 
\begin{itemize}
    \item \textbf{roach.mp4} is an \emph{uncut} evaluation run recorded from Roach driving in Town03 (LeaderBoard-busy under dynamic weather). This video demonstrates the natural driving style of Roach even in challenging situations such as US-style traffic lights, unprotected left turns, roundabouts and stop signs.
    \item \textbf{autopilot.mp4} is an \emph{uncut} evaluation run recorded from Autopilot driving in Town02 (NoCrash-dense, new town \& new weather). This video demonstrates the over-conservative behavior of the Autopilot while driving in dense traffic. This often leads to red light infractions and blockage (both are present in the video).
    \item \textbf{il\_agent.mp4} is a \emph{highlight} video recorded from our best roach-supervised IL agent $\mathcal{L}_\text{K}+\mathcal{L}_\text{F}(\text{c})$. This video includes multiple challenging situations often encountered during urban driving, such as EU and US-style junctions, unprotected left turns, roundabouts and reacting to pedestrians walking into the street. Furthermore, we highlight some of the failure modes of our camera-based IL agent, including not coming to a full stop for stop signs, collisions at overcrowded intersections and oscillation in the steering if the lane markings are not visible due to sun glare. We believe that including memory in the IL agent policy can help in most of these issues, due to a better understanding of the ego-motion (stop sign and oscillations) and other agents' motion (collisions). 
\end{itemize}

\subsection{Code}
To reproduce our results, we provide four python scripts:
\begin{itemize}
    \item \emph{train\_rl.py} for training Roach.
    \item \emph{train\_il.py} for training DA-RB (CILRS + DAGGER).
    \item \emph{benchmark.py} for benchmarking agents.
    \item \emph{data\_collect.py} for collecting on/off-policy data.
\end{itemize}

It is recommended to run our scripts through bash files contained in the folder \emph{run}. 
All configurations are in the folder \emph{config}. 
Our repository is composed of two modules:
\begin{itemize}
    \item \emph{carla\_gym}, a versatile OpenAI gym \cite{OpenaiGym} environment for CARLA. It allows not only RL training with synchronized rollouts, but also data collection and evaluation. The environment is configurable in terms of weather, number of background pedestrians and vehicles, benchmarks, terminal conditions, sensors, rewards for the ego-vehicle and etc.
    \item \emph{agents}, which includes our implementation of Autopilot (in \emph{agents/expert}), Roach (in \emph{agents/rl\_birdview}) and DA-RB (in \emph{agents/cilrs}).
\end{itemize}

\subsection{Rendering issues}
\label{sec: rendering}
\begin{figure*}[t]
     \centering
     \begin{subfigure}[b]{0.96\textwidth}
         \centering
         \includegraphics[width=\textwidth]{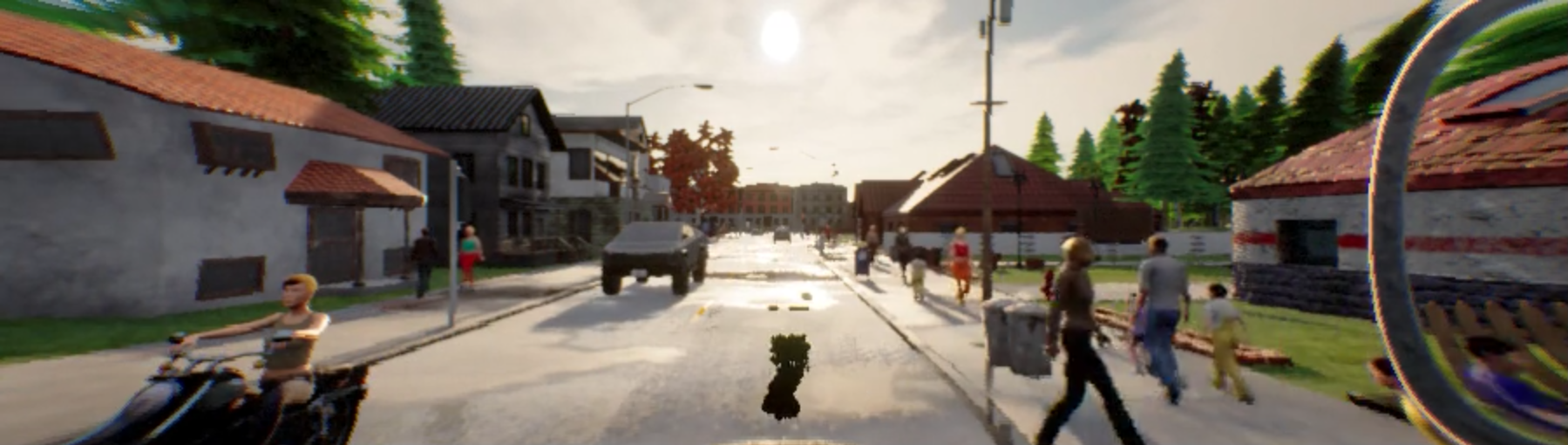}
         \caption{Reflections from after-rain puddles in fornt of the ego-vehicle are incorrectly rendered as black pixels.}
         \label{fig:reflection_bad}
     \end{subfigure}
     \begin{subfigure}[b]{0.96\textwidth}
         \centering
         \vspace{1ex}
         \includegraphics[width=\textwidth]{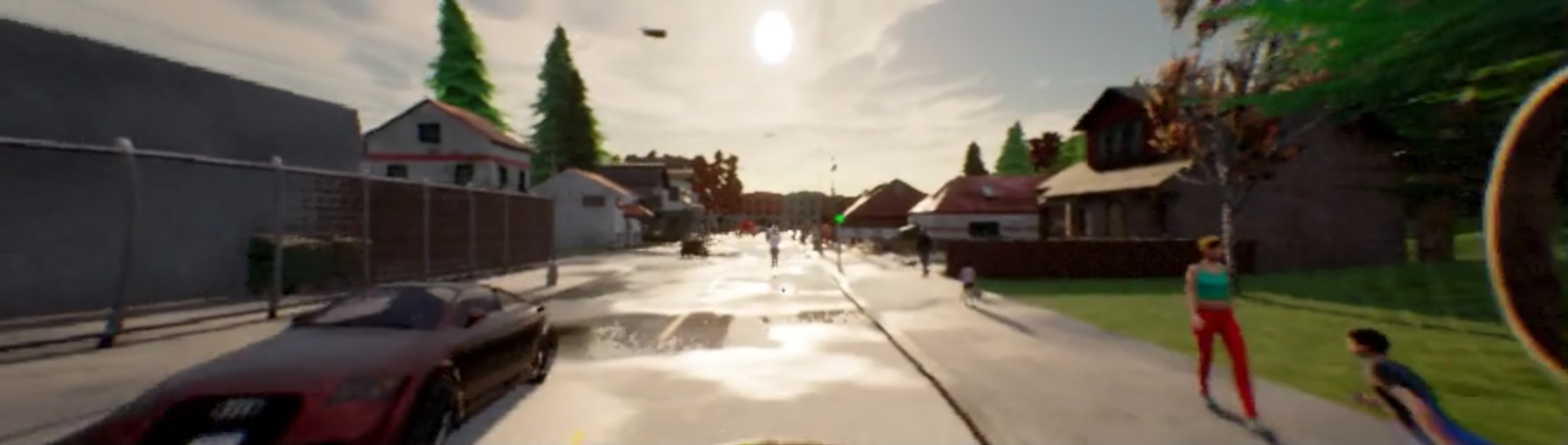}
         \caption{Reflections are correctly rendered if the puddle is not directly in front of the ego-vehicle.}
         \label{fig:reflection_good}
     \end{subfigure}
    \vspace{-1ex}
    \caption{\textbf{Rendering issue of CARLA 0.9.11 running on Ubuntu with OpenGL.}}
    \vspace{-2ex}
    \label{fig:reflection}
\end{figure*}

As illustrated in Fig.~\ref{fig:reflection}, on CARLA 0.9.11 reflections from after-rain puddles are sometimes wrongly rendered as black pixels.
When the black pixels are accumulated, for example in the middle of Fig.~\ref{fig:reflection_bad}, they are often recognized as obstacles by the camera-based agents.
Since this kind of reflection only appears under the testing weather but not under the training weather, generalizing to testing weather is exceptionally hard on CARLA 0.9.11 for the camera-based end-to-end IL agents.

\section{Implementation Details}

\subsection{Roach}
The network architecture of Roach can be found in Table~\ref{table:roach_architecture} and the hyper-parameter values are listed in Table~\ref{table:roach_parameter}.

\vspace{1ex}\noindent{\bf BEV:}
Cyclists and pedestrians are rendered larger than their actual sizes, this allows us to use a smaller image encoder with less parameters for Roach. Additionally, increasing the size naturally adds some caution when dealing with these vulnerable road users. 

\vspace{1ex}\noindent{\bf Update:} 
The policy network and the value network are updated together using one Adam optimizer with an initial learning rate of 1e-5.
The learning rate is scheduled based on the empirical KL-divergence between the policy before and after the update. 
If the KL-divergence is too large after an update epoch, the update phase will be interrupted and a new rollout phase will start. 
Furthermore, a patience counter will be increased by one and the learning rate will be reduced once the patience counter reaches a threshold.

\vspace{1ex}\noindent{\bf Rollout:}
Before each update phase a fixed-size buffer will be filled with trajectories collected on six CARLA servers, each corresponds to one of the six LeaderBoard maps (Town1-6).

\vspace{1ex}\noindent{\bf Terminal Condition:}
An episode is terminated if and only if one of the following event happens.
\begin{itemize}
    \item Run red light: examination code taken from the public repository of LeaderBoard. Terminal reward: $-1-s$.
    \item Run stop sign: examination code taken from the public repository of LeaderBoard. Terminal reward: $-1-s$.
    \item Collision registered by CARLA: based on the physics engine. Any collision with intensity larger than 0 is considered. Terminal reward: $-1-s$.
    \item Collision detected by bounding box overlapping in the BEV. Terminal reward: $-1-s$.
    \item Route deviation: triggered if the lateral distance to the lane centerline of the desired route is larger than 3.5 meters. Terminal reward: $-1$.
    \item Blocked: speed of the ego-vehicle is slower than 0.1 m/s for more than 90 consecutive seconds. Terminal reward: $-1$.
\end{itemize}
with $s$ is the ego-vehicle's speed.
The terminal reward is the reward given to the very last observation/action pair before the termination.
For non-terminal samples, the terminal reward is 0.

\vspace{1ex}\noindent{\bf Reward Shaping:}
The reward is the sum of the following components.
\begin{itemize}
    \item r\_speed: equals to $1.0 - |s - s_\text{desired}|/s_\text{max}$, where $s$ is the measured speed of the ego-vehicle, $s_\text{max}$ is the maximum speed and $s_\text{desired}$ is the desired speed. We use a constant maximum speed $s_\text{max}=6$ m/s. The desired speed is a variable and is explained below.
    \item r\_position: equals to $-0.5\Delta_\text{p}$, where $\Delta_\text{p}$ is the lateral distance (in meters) between the ego-vehicle's center and the center line of the desired route.
    \item r\_rotation: equals to $-\Delta_\text{r}$, where $\Delta_\text{r}$ is the absolute value of the angular difference (in radians) between the ego-vehicle's heading and the heading of the center line of the desired route.
    \item r\_action: equals to $-0.1$ if the current steering differs more than 0.01 from the steering applied in the previous step.
    \item r\_terminal: the aforementioned terminal reward.
\end{itemize}
The desired speed, as proposed in \cite{toromanoff2019deep}, depends on rule-based obstacle detections.
If there's no obstacle detected, the desired speed equals to the maximum speed.
If an obstacle is detected, based on the distance to the obstacle the desired speed is linearly decreased to 0.
As obstacle detector we use the hazard detection of Autopilot (cf. Section \ref{sec:autopilot}).
As a dense and informative reward, r\_speed helps substantially to train our Roach and the camera-based end-to-end RL agent \cite{toromanoff2019deep}.
However, using rule-based obstacle detections inevitable introduces bias, the trained RL agent can be over-aggressive or over-conservative depending on the false positive and false negative rate of the detector.
For example, during multi-lane freeway driving, our Roach decelerates for vehicles on the neighbouring lanes because those vehicles are detected as obstacles during training.
Another example, Roach tends to collide after a right turn, this is related to the sector shaped (around 40 degrees) detection area used by the obstacle detection; vehicles and pedestrians on the right are not covered in the detection area.
To further improve the performance of Roach, this r\_speed should be modified, either using a better obstacle detector, or completely remove the rule-based obstacle detection, and build a less artificial reward based on simulation states.

\vspace{1ex}\noindent{\bf Mode of Beta Distribution:}
We take the distribution mode as a deterministic output.
The mode of the Beta distribution $\mathcal{B}(\alpha,\beta)$ is defined as
\begin{equation}
  \text{M}  =
    \begin{cases}
      \frac{\alpha-1}{\alpha+\beta-2} & \text{if $\alpha>1, \beta>1$}\\
      0 & \text{if $\alpha\leq1, \beta>1$}\\
      1 & \text{if $\alpha>1, \beta\leq1$}\\
      \text{bimodal $\{0,1\}$} & \text{if $\alpha<1, \beta<1$}\\
      \text{any value in $[0,1]$} & \text{if $\alpha=1, \beta=1$}
    \end{cases}       
\end{equation}
For a natural driving behavior, we use the mean $\frac{\alpha}{\alpha+\beta}$ as the deterministic output when the mode is not uniquely defined, i.e. when $\alpha<1, \beta<1$ or $\alpha=1, \beta=1$.

\subsection{IL Agent Supervised by Roach}

The network architecture of our IL agent is found in Table~\ref{table:cilrs_architecture} and the hyper-parameter values are listed in Table~\ref{table:cilrs_parameter}.

\vspace{1ex}\noindent{\bf Network Architecture:}
We use six branches: turning left, turning right and going straight at the junction, following lane, changing to the left lane and changing to the right lane.

\vspace{1ex}\noindent{\bf Off-policy Data Collection:}
Following CILRS \cite{codevilla2019exploring}, triangular perturbations on actions are applied while collecting the off-policy expert dataset to alleviate the covariate shift.
The off-policy dataset for NoCrash includes 80 episodes and for LeaderBoard it includes 160 episodes.
Each episode is at most 300 seconds and at least 30 seconds long.
The episode will be terminated if the expert violates any traffic rules, including red light infractions, stop sign infractions and collisions.
In such a case, we remove the last 30 seconds of that episode so as to ensure that the off-policy dataset includes only correct demonstrations.
Data is not collected using the given training routes but from randomly spawned start and target locations.

\vspace{1ex}\noindent{\bf On-policy Data Collection:}
We follow DA-RB \cite{prakash2020exploring} for DAGGER with critical state sampling and replay buffer. 
New DAGGER-data will replace the old data in the replay buffer, while the buffer size is fixed.
The same number of frames are contained in the replay buffer as in the off-policy dataset.
At each DAGGER iteration, around 15-25\% of the replay buffer is filled with new DAGGER-data, whereas at least 20\% of the replay buffer is filled with off-policy data.
Identical to the off-policy data collection, we use randomly spawned start and target locations while collecting DAGGER datasets.
Following DA-RB, we did not use a mixed agent/expert policy to collect DAGGER datasets.
However, our code allows this kind of rollout for DAGGER.

\vspace{1ex}\noindent{\bf Training Details:}
Since we take the ResNet-34 pre-trained on ImageNet, the input image is normalized as suggested.
In case the IL agent uses a distributional action head and/or a value head, the corresponding weights will be loaded from the Roach model at the first training iteration (the behavior cloning iteration).
At each DAGGER iteration, the training continuous from the last epoch of the previous DAGGER iteration.
We apply image augmentations using code modified from CILRS. 
The image augmentation methods are applied in random order and include Gaussian blur, additive Gaussian noise, coarse and block-wise dropouts, additive and multiplicative noise to each channel, randomized contrast and grayscale.
All models are trained for 25 epochs using the ADAM optimizer with an initial learning rate of 2e-4. 
The learning rate is halved if the validation loss has not decreased for more than 5 epochs.

\subsection{Autopilot}
\label{sec:autopilot}
The CARLA Autopilot (also called roaming agent) is a simple but effective automated expert based on hand-crafted rules and ground-truth simulation states. The Autopilot is composed of two PID controllers for trajectory tracking and hazard detectors for emergency brake. Hazards include
\begin{itemize}
    \item pedestrians/vehicles detected ahead,
    \item red lights/stop sings detected ahead,
    \item negative ego-vehicle speed, for handling slopes.
\end{itemize}
Locations and states of pedestrians, vehicles, red lights and stop signs are provided as ground-truth by the CARLA API. If any hazard appears in a trigger area ahead of the ego-vehicle, Autopilot will make an emergency brake with $throttle=0,$ $steering=0$, $brake=1$.
If no hazard is detected, the ego-vehicle will follow the desired path using two PID controllers, one for speed and one for steering control. The PID controller takes as input the location, rotation and speed of the ego-vehicle and the desired route specified as dense (1 meter interval) waypoints. The speed PID yields $throttle\in[0,1]$ and the steering PID yields $steering\in[-1,1]$. We tuned the parameters for PID controllers and hazard detectors manually, such that the Autopilot is a strong baseline. The target speed is 6 m/s.

\section{Benchmarks}
\noindent{\bf Scope:}
The scope of the NoCrash and the offline LeaderBoard benchmark are illustrated in Table~\ref{table:benchmark_scope}.
The offline LeaderBoard benchmark considers more traffic scenarios and longer routes in six different maps.


\vspace{1ex}\noindent{\bf Weather:}
Following the NoCrash benchmark, we use \textit{ClearNoon}, \textit{WetNoon}, \textit{HardRainNoon} and \textit{ClearSunset} as the training weather types, whereas new weather types are \textit{SoftRainSunset} and \textit{WetSunset}.
To save computational resources, only two out of the four training weather types are evaluated, they are \textit{WetNoon} and \textit{ClearSunset}.

\vspace{1ex}\noindent{\bf Background Traffic:}
The number of vehicles and pedestrians spawned in each map of different benchmarks are listed in Table~\ref{table:benchmark_traffic}. 
Vehicles and pedestrians are spawned randomly from the complete blueprint library of CARLA 0.9.11. This stands in contrast to several previous works where for example two-wheeled vehicles are disabled.

\vspace{1ex}\noindent{\bf Pros and cons of the online and the offline Leaderboard:}

Online Leaderboard:
(+) All methods are evaluated under exactly the same condition.
(+) No need to re-evaluate other methods.
(-) No restriction on how the method is trained and how the training data is collected. 

Offline Leaderboard:
(+) Strictly prescribes both the training and testing environment.
(+) Full control and observation over the benchmark.
(-) You will have to re-evaluate other methods, if any setup of the benchmark has changed, for example CARLA version and etc.

One can use the offline Leaderboard if a thorough study on the generalization ability of the method is desired.

\section{Additional Experimental Results}

\begin{figure}[t]
\begin{center}
    \includegraphics[width=\linewidth]{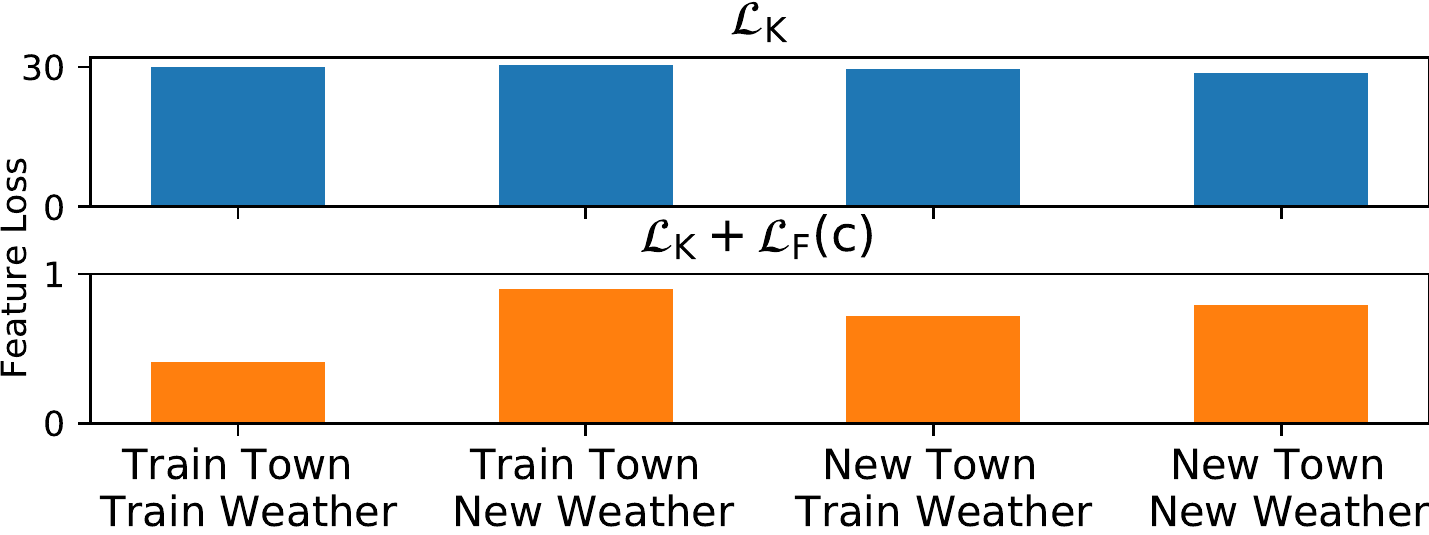}
\end{center}
\vspace{-3ex}
\caption{\textbf{Feature loss w.r.t. Roach} on one of the NoCrash-dense route. The y-axis of both charts have different scale.}
\vspace{-1ex}
\label{fig:feature_loss}
\end{figure}

To verify IL agents trained using the feature loss indeed embed camera images to the latent space of Roach, we report the feature loss at test time in Fig. \ref{fig:feature_loss}.
In the first row of Fig. \ref{fig:feature_loss}, the IL agent trained without feature loss, $\mathcal{L}_\text{K}$, learns a latent space independent of the one of Roach.
Hence, the test feature loss is effectively noise that is invariant to the test condition.
In the second row, $\mathcal{L}_\text{K}+\mathcal{L}_\text{F}(\text{c})$ is trained with the feature loss.
The test feature loss of this agent is much smaller (less than 1) and increases as expected during the generalization tests.

\begin{table}
\setlength{\tabcolsep}{9pt}
\centering
\begin{tabular}{lcccc}
\toprule
\textbf{Map}
& \begin{tabular}{@{}c@{}} \textbf{\#} \\ \textbf{Routes} \end{tabular} 
& \begin{tabular}{@{}c@{}} \textbf{Total} \\ \textbf{Km} \end{tabular}
& \begin{tabular}{@{}c@{}} \textbf{\# Traffic} \\ \textbf{lights} \end{tabular} 
& \begin{tabular}{@{}c@{}} \textbf{\# Stop} \\ \textbf{signs} \end{tabular}
\\ 
\cmidrule(lr){1-5}
\multicolumn{5}{c}{NoCrash Train}  \\ 
\cmidrule(lr){1-5}
Town01 & $25$ & $17.4$ & $110$ & $0$ \\
\cmidrule(lr){1-5}
\multicolumn{5}{c}{NoCrash Test}  \\ 
\cmidrule(lr){1-5}
Town02 & $25$ & $8.9$ & $94$ & $0$ \\
\cmidrule(lr){1-5}
\multicolumn{5}{c}{LeaderBoard Train}  \\ 
\cmidrule(lr){1-5}
Town01 & $10$ & $7.9$ & $47$ & $0$ \\
Town03 & $20$ & $30.7$ & $140$ & $63$ \\
Town04 & $10$ & $24.1$ & $72$ & $13$ \\
Town06 & $10$ & $19.5$ & $58$ & $1$ \\
\cmidrule(lr){1-5}
\multicolumn{5}{c}{LeaderBoard Test}  \\ 
\cmidrule(lr){1-5}
Town02 & $6$ & $5.5$ & $54$ & $0$ \\
Town04 & $10$ & $24.1$ & $72$ & $13$ \\
Town05 & $10$ & $12.4$ & $82$ & $29$ \\
\bottomrule
\end{tabular}
\vspace{-1ex}
\caption{\textbf{Scope of the Nocrash benchmark and the offline LeaderBoard benchmark.} Total kilometers, number of traffic lights and stop signs are measured using Roach.}
\vspace{-2ex}
\label{table:benchmark_scope}
\end{table}

\begin{figure*}[t]
    \centering
     \begin{subfigure}[b]{\textwidth}
         \centering
         \includegraphics[width=\textwidth]{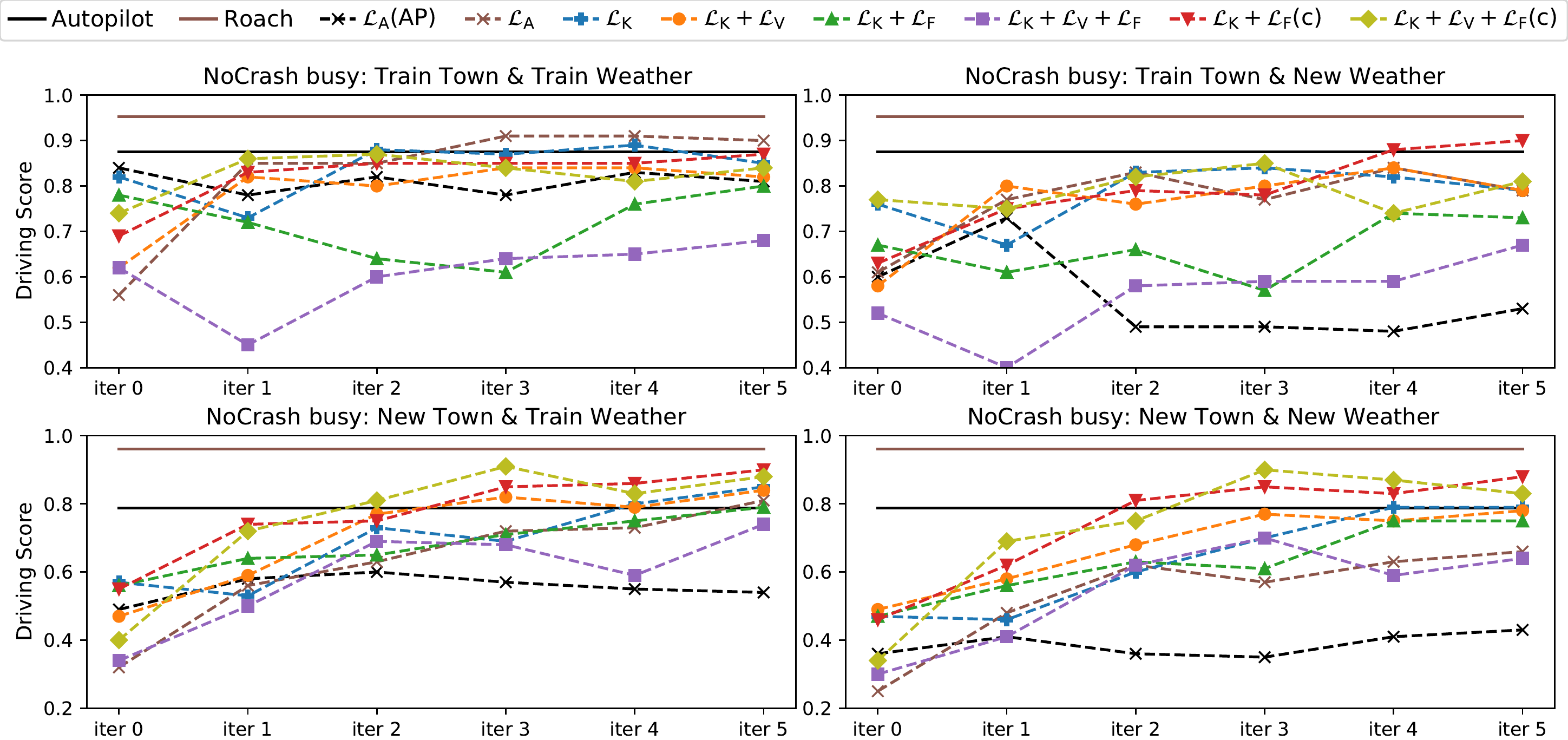}
         \caption{\textbf{Driving Score}}
         \vspace{3ex}
     \end{subfigure}
     \begin{subfigure}[b]{\textwidth}
         \centering
         \includegraphics[width=\textwidth]{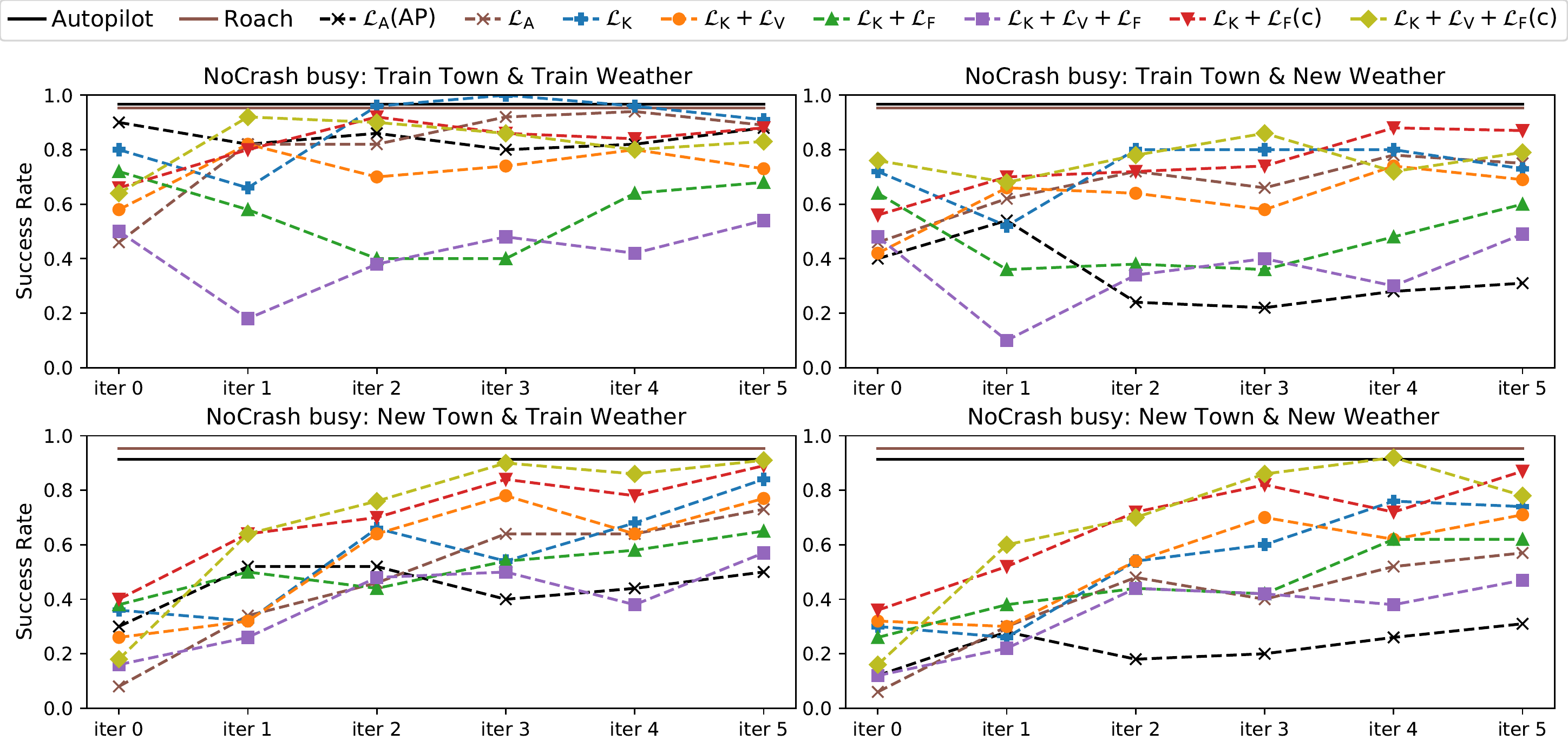}
         \caption{\textbf{Success Rate}}
     \end{subfigure}
    \caption{\textbf{Driving performance of experts and IL agents on the NoCrash-busy benchmark.} All IL agents (dashed lines) are supervised by Roach except for $\mathcal{L}_\text{A}(\text{AP})$, which is supervised by the CARLA Autopilot. For IL agents at the 5th iteration and all experts, results are reported as the mean over 3 evaluation seeds. Others agents are evaluated only once.}
\label{fig:performance_eu}
\end{figure*}

\begin{figure*}[t]
    \centering
     \begin{subfigure}[b]{\textwidth}
         \centering
         \includegraphics[width=\textwidth]{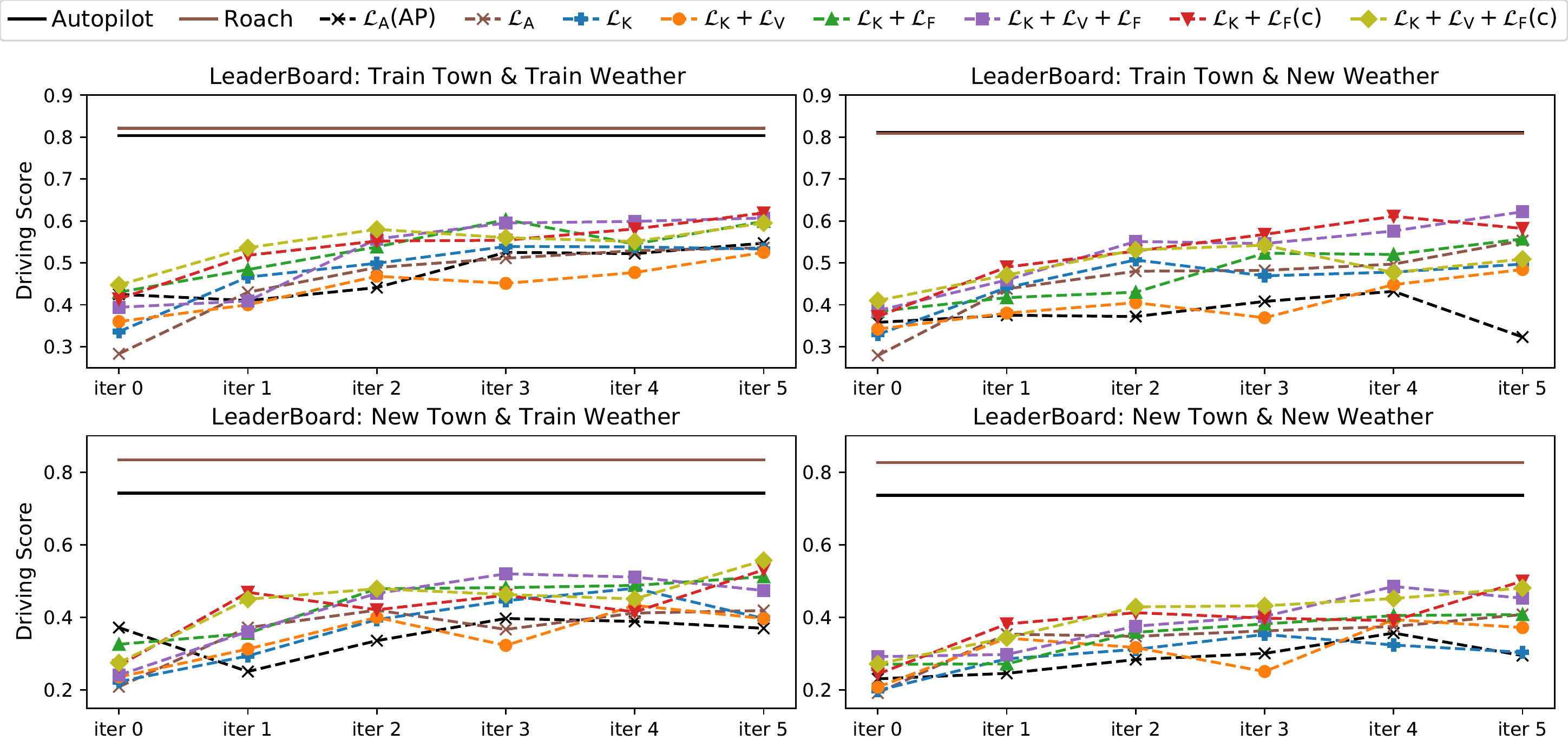}
         \caption{\textbf{Driving Score}}
         \vspace{3ex}
     \end{subfigure}
     \begin{subfigure}[b]{\textwidth}
         \centering
         \includegraphics[width=\textwidth]{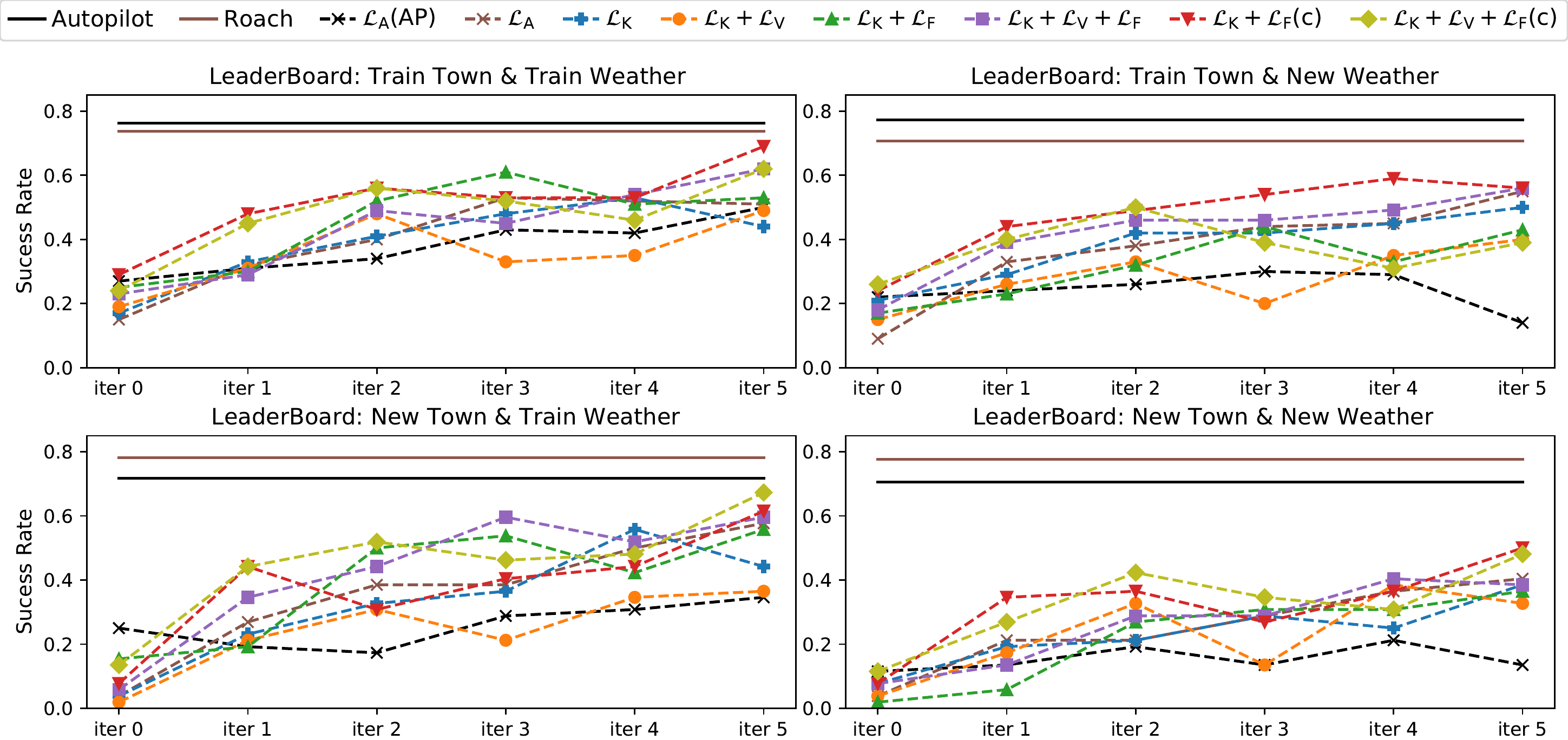}
         \caption{\textbf{Success Rate}}
     \end{subfigure}
    \caption{\textbf{Driving performance of experts and IL agents on the offline LeaderBoard-busy benchmark.} All IL agents (dashed lines) are supervised by Roach except for $\mathcal{L}_\text{A}(\text{AP})$, which is supervised by the CARLA Autopilot. For all experts, results are reported as the mean over 3 evaluation seeds. Results of IL agents are evaluated only once.}
\label{fig:performance_lb}
\end{figure*}

To complete Fig. 5 of the main paper, driving scores of experts and IL agents at each DAGGER iterations are in Fig.~\ref{fig:performance_eu} (NoCrash-busy) and Fig.~\ref{fig:performance_lb} (LeaderBoard-busy).

To complete Table 3 of the main paper, detailed driving performance and infraction analysis of our experts and IL agents (5th DAGGER iteration) are listed in
\begin{itemize}
    \item Table~\ref{table:infraction_eu_tt}: NoCrash-busy, train town \& train weather.
    Table~\ref{table:infraction_eu_tn}: NoCrash-busy, train town \& new weather.
    Table~\ref{table:infraction_eu_nt}: NoCrash-busy, new town \& train weather.
    Table~\ref{table:infraction_eu_nn}: NoCrash-busy, new town \& new weather.
    \item Table~\ref{table:infraction_lb_tt}: LeaderBoard, train town \& train weather. 
    Table~\ref{table:infraction_lb_tn}: LeaderBoard, train town \& new weather. \\
    Table~\ref{table:infraction_lb_nt}: LeaderBoard, new town \& train weather. \\
    Table~\ref{table:infraction_lb_nn}: LeaderBoard, new town \& new weather.
\end{itemize}
\clearpage

\begin{table*}
\setlength{\tabcolsep}{12pt}
\centering
\begin{tabular}{lcc}
\toprule
\textbf{Map}
& \textbf{\# Vehicles}
& \textbf{\# Pedestrians}
\\ 
\cmidrule(lr){1-3}
\multicolumn{3}{c}{NoCrash dense}  \\ 
\cmidrule(lr){1-3}
Town01 & $100$ & $250$ \\
Town02 & $70$ & $150$ \\
\cmidrule(lr){1-3}
\multicolumn{3}{c}{NoCrash busy}  \\ 
\cmidrule(lr){1-3}
Town01 & $120$ & $120$ \\
Town02 & $70$ & $70$ \\
\cmidrule(lr){1-3}
\multicolumn{3}{c}{LeaderBoard busy}  \\ 
\cmidrule(lr){1-3}
Town01 & $120$ & $120$ \\
Town02 & $70$ & $70$ \\
Town03 & $70$ & $70$ \\
Town04 & $150$ & $80$ \\
Town05 & $120$ & $120$ \\
Town06 & $120$ & $80$ \\
\bottomrule
\end{tabular}
\caption{\textbf{Background traffic settings for different benchmarks.}}
\label{table:benchmark_traffic}
\end{table*}
\begin{table*}
\setlength{\tabcolsep}{5.8pt}
\centering
\begin{tabular}{lcccc}
\toprule
\textbf{Layer Type} & \textbf{Filters} & \textbf{Size} & \textbf{Strides} & \textbf{Activation} \\
\cmidrule(lr){1-5}
\multicolumn{5}{c}{Image Encoder}  \\ 
\cmidrule(lr){1-5}
Conv2d & 8 & 5x5 & 2 & ReLU \\ 
Conv2d & 16 & 5x5 & 2 & ReLU \\ 
Conv2d & 32 & 5x5 & 2 & ReLU \\ 
Conv2d & 64 & 3x3 & 2 & ReLU \\ 
Conv2d & 128 & 3x3 & 2 & ReLU \\ 
Conv2d & 256 & 3x3 & 1 & - \\
Flatten & & & & \\
\cmidrule(lr){1-5}
\multicolumn{5}{c}{Measurement Encoder}  \\ 
\cmidrule(lr){1-5}
Dense & 256 & & & ReLU \\
Dense & 256 & & & ReLU \\
\cmidrule(lr){1-5}
\multicolumn{5}{c}{FC Layers after Concatenation}  \\ 
\cmidrule(lr){1-5}
Dense & 512 & & & ReLU \\
Dense & 256 & & & ReLU \\
\cmidrule(lr){1-5}
\multicolumn{5}{c}{Action Head}  \\ 
\cmidrule(lr){1-5}
Dense (shared) & 256 & & & ReLU \\
Dense (shared) & 256 & & & ReLU \\
Dense (for $\alpha$) & 2 & & & Softplus \\
Dense (for $\beta$) & 2 & & & Softplus \\
\cmidrule(lr){1-5}
\multicolumn{5}{c}{Value Head}  \\ 
\cmidrule(lr){1-5}
Dense & 256 & & & ReLU \\
Dense & 256 & & & ReLU \\
Dense & 1 & & & - \\
\bottomrule
\end{tabular}
\caption{\textbf{The network architecture used for Roach.} Around 1.53M trainable parameters.}
\label{table:roach_architecture}
\end{table*}
\begin{table*}
\setlength{\tabcolsep}{8.6pt}
\centering
\begin{tabular}{lccc}
\toprule
\textbf{Layer Type} & \textbf{Filters} & \textbf{Activation} & \textbf{Dropout} \\
\cmidrule(lr){1-4}
\multicolumn{4}{c}{Image Encoder}  \\ 
\cmidrule(lr){1-4}
ResNet-34 &  &  &   \\
\cmidrule(lr){1-4}
\multicolumn{4}{c}{Measurement Encoder}  \\ 
\cmidrule(lr){1-4}
Dense & 128  & ReLU &  \\
Dense & 128  & ReLU &  \\
\cmidrule(lr){1-4}
\multicolumn{4}{c}{FC Layers after concatenation}  \\ 
\cmidrule(lr){1-4}
Dense & 512 & ReLU & \\
Dense & 512 & ReLU & \\
Dense & 256 & ReLU & \\
\cmidrule(lr){1-4}
\multicolumn{4}{c}{Speed Head}  \\ 
\cmidrule(lr){1-4}
Dense & 256 & ReLU &  \\
Dense & 256 & ReLU & 0.5\\
Dense & 1 & & \\
\cmidrule(lr){1-4}
\multicolumn{4}{c}{Value Head}  \\ 
\cmidrule(lr){1-4}
Dense & 256 & ReLU &  \\
Dense & 256 & ReLU & 0.5\\
Dense & 1 & & \\
\cmidrule(lr){1-4}
\multicolumn{4}{c}{Deterministic Action Head}  \\ 
\cmidrule(lr){1-4}
Dense & 256 & ReLU &  \\
Dense & 256 & ReLU & 0.5\\
Dense & 2 & & \\
\cmidrule(lr){1-4}
\multicolumn{4}{c}{Distributional Action Head}  \\ 
\cmidrule(lr){1-4}
Dense (shared) & 256 & ReLU &  \\
Dense (shared) & 256 & ReLU & 0.5\\
Dense (for $\alpha$) & 2 & Softplus \\
Dense (for $\beta$) & 2 & Softplus \\
\bottomrule
\end{tabular}
\caption{\textbf{The network architecture used for our IL agent.} Around 23.4M trainable parameters.}
\label{table:cilrs_architecture}
\end{table*}
\begin{table*}
\setlength{\tabcolsep}{10pt}
\centering
\begin{tabular}{llc}
\toprule
\textbf{Notation} & \textbf{Description} & \textbf{Value}  \\ 
\cmidrule(lr){1-3}
\multicolumn{3}{c}{BEV Representation} \\
\cmidrule(lr){1-3}
$W$ & Width & 192 px \\
$H$ & Height & 192 px \\
$C$ & Number of channels & 15   \\
$K$ & Size of the temporal sequence & 4 \\
 & Timestamps of images in the temporal sequence & \{-1.5, -1, -0.5, 0\} sec \\
$D$ & Distance from the ego-vehicle to the bottom & 40 px  \\
& Pixels per meter & 5 px/m  \\
& Minimum width/height of rendered bounding boxes & 8 px \\
& Scale factor for bounding box size of pedestrians & 2 \\
\cmidrule(lr){1-3}
\multicolumn{3}{c}{Rollout} \\
\cmidrule(lr){1-3}
& Buffer size for six environments & 12288 frames \\
& Value bootstrap for the last non-terminal sample & True \\
& Synchronized & True \\
& Reset at the beginning of a new phase & False \\
& Weather & dynamic \\
& Range of vehicle/pedestrian number in Town 1 & $[0,150]/[0,300]$ \\
& Range of vehicle/pedestrian number in Town 2 & $[0,100]/[0,200]$ \\
& Range of vehicle/pedestrian number in Town 3 & $[0,120]/[0,120]$ \\
& Range of vehicle/pedestrian number in Town 4 & $[0,160]/[0,160]$ \\
& Range of vehicle/pedestrian number in Town 5 & $[0,160]/[0,160]$ \\
& Range of vehicle/pedestrian number in Town 6 & $[0,160]/[0,160]$ \\
 \cmidrule(lr){1-3}
\multicolumn{3}{c}{Update} \\
\cmidrule(lr){1-3}
 & Number of epochs & 20 \\
$\lambda_\text{ent}$ & Weight for the entropy loss & 0.01 \\
$\lambda_\text{exp}$ & Weight for the exploration loss & 0.05 \\
& Weight for value loss & 0.5 \\
& $\gamma$ for GAE & 0.99\\
& $\lambda$ for GAE & 0.9 \\
& Clipping range for PPO-clip & 0.2 \\
& Max norm for gradient clipping & 0.5\\
& Batch size & 256 \\
& Initial learning rate & 1e-5 \\
& KL-divergence threshold for learning rate schedule & 0.15 \\
& Patience for learning rate schedule & 8 \\
& Factor for learning rate schedule & 0.5 \\
\bottomrule
\end{tabular}
\caption{\textbf{The hyper-parameter values used for Roach.}}
\label{table:roach_parameter}
\end{table*}
\begin{table*}
\setlength{\tabcolsep}{10pt}
\centering
\begin{tabular}{lc}
\toprule
 \textbf{Description} & \textbf{Value}  \\ 
\cmidrule(lr){1-2}
\multicolumn{2}{c}{Inputs} \\
\cmidrule(lr){1-2}
Camera type & RGB \\
Camera image width & 900 px \\
Camera image height & 256 px \\
Camera location $[x,y,z]$ relative to the ego-vehicle & $[-1.5, 0, 2]$ \\
Camera rotation $[roll,pitch,yaw]$ relative to the ego-vehicle & $[0,0,0]$ \\
Camera horizontal FOV & $100^{\circ}$ \\
Mean for image normalization & $[0.485, 0.456, 0.406]$ \\
Standard deviation for image normalization & $[0.229, 0.224, 0.225]$ \\
Speed measurement & Forward speed in m/s \\
Normalization factor for speed & 12 \\
\cmidrule(lr){1-2}
\multicolumn{2}{c}{Data Collection} \\
\cmidrule(lr){1-2}
Episode length & 300 sec \\
Triangular perturbation for off-policy data & 20\% \\
Number of episodes (NoCrash, off-policy) & 80 \\
Number of episodes (LeaderBoard, off-policy) & 160 \\
Number of episodes (NoCrash, on-policy, Autopilot) & 80 \\
Number of episodes (LeaderBoard, on-policy, Autopilot) & 160 \\
Number of episodes (NoCrash, on-policy, Roach) & 40 \\
Number of episodes (LeaderBoard, on-policy, Roach) & 80 \\
DA-RB critical state sampling criterion & difference in acceleration \\
DA-RB critical state sampling threshold & 0.2 \\
Weather & Same as NoCrash train weathers \\
Range of vehicle/pedestrian number in NoCrash train town 1  & $[0,150]/[0,200]$ \\
Range of vehicle/pedestrian number in LeaderBoard train town 1  & $[80,160]/[80,160]$ \\
Range of vehicle/pedestrian number in LeaderBoard train town 3 & $[40,100]/[40,100]$ \\
Range of vehicle/pedestrian number in LeaderBoard train town 4 & $[100,200]/[40,120]$ \\
Range of vehicle/pedestrian number in LeaderBoard train town 6 & $[80,160]/[40,120]$ \\
 \cmidrule(lr){1-2}
\multicolumn{2}{c}{Training} \\
\cmidrule(lr){1-2}
Number of epochs at each DAGGER iteration & 25 \\
$\lambda_\text{S}$, weight for the speed regularization & 0.05 \\
$\lambda_\text{V}$, weight for the value loss, if applied & 0.05 \\
$\lambda_\text{F}$, weight for the feature loss, if applied & 0.001 \\
Batch size & 48 \\
Initial learning rate & 0.0002 \\
Patience for reduce-on-plateau learning rate schedule & 5 \\
Factor for learning rate schedule & 0.5 \\
Pre-trained distributional action head & True \\
Pre-trained value head & True \\
Image augmentation & True \\
\bottomrule
\end{tabular}
\caption{\textbf{The hyper-parameter values used for our IL agent.}}
\label{table:cilrs_parameter}
\end{table*}


\begin{table*}
\setlength{\tabcolsep}{2.6pt}
\centering
\begin{tabular}{lccccccccc} 
\toprule
& \begin{tabular}{@{}c@{}}Success \\ rate \end{tabular} 
& \begin{tabular}{@{}c@{}}Driving \\ score \end{tabular} 
& \begin{tabular}{@{}c@{}}Route \\ compl. \end{tabular} 
& \begin{tabular}{@{}c@{}}Infrac. \\ penalty \end{tabular} 
& \begin{tabular}{@{}c@{}}Collision \\ others \end{tabular} 
& \begin{tabular}{@{}c@{}}Collision \\ pedestrian \end{tabular} 
& \begin{tabular}{@{}c@{}}Collision \\ vehicle \end{tabular}  
& \begin{tabular}{@{}c@{}}Red light \\ infraction \end{tabular}  
& \begin{tabular}{@{}c@{}}Agent \\ blocked \end{tabular}  \\
\cmidrule(lr){1-1}\cmidrule(lr){2-5}\cmidrule(lr){6-10}
iter 5
& \%, $\uparrow$
& \%, $\uparrow$
& \%, $\uparrow$
& \%, $\uparrow$
& \#/Km, $\downarrow$
& \#/Km, $\downarrow$
& \#/Km, $\downarrow$
& \#/Km, $\downarrow$
& \#/Km, $\downarrow$
\\
\cmidrule(lr){1-1}\cmidrule(lr){2-5}\cmidrule(lr){6-10}
$\mathcal{L}_\text{A}(\text{AP})$
& $88\pm2$ & $81\pm2$ & $94\pm2$ & $86\pm1$ 
& $\mathbf{0}\pm0$ & $\mathbf{0}\pm0$ & $0.08\pm0.11$ & $1.02\pm0.33$ & $1\pm0.28$ \\
$\mathcal{L}_\text{A}$
& $89\pm5$ & $\mathbf{90}\pm2$ & $\mathbf{99}\pm1$ & $90\pm1$ 

& $0.06\pm0.04$ & $0.05\pm0.02$ & $0.06\pm0.04$ & $0.29\pm0.03$ & $\mathbf{0.05}\pm0.06$ \\
$\mathcal{L}_\text{K}$
& $\mathbf{91}\pm10$ & $85\pm6$ & $99\pm2$ & $85\pm5$ 
& $0.1\pm0.18$ & $0.03\pm0.04$ & $0.1\pm0.11$ & $0.58\pm0.07$ & $0.07\pm0.12$ \\
$\mathcal{L}_\text{K}+\mathcal{L}_\text{V}$
& $73\pm4$ & $82\pm3$ & $91\pm2$ & $\mathbf{91}\pm2$
& $0.07\pm0.07$ & $0.02\pm0.02$ & $0.18\pm0.12$ & $\mathbf{0.27}\pm0.06$ & $0.6\pm0.2$ \\
$\mathcal{L}_\text{K}+\mathcal{L}_\text{F}$
& $68\pm11$ & $80\pm6$ & $89\pm3$ & $89\pm4$ 
& $0.15\pm0.03$ & $0.02\pm0.01$ & $\mathbf{0.05}\pm0.06$ & $0.41\pm0.13$ & $0.12\pm0.02$ \\
$\mathcal{L}_\text{K}+\mathcal{L}_\text{V}+\mathcal{L}_\text{F}$
& $54\pm2$ & $68\pm3$ & $80\pm2$ & $87\pm3$ 
& $0.22\pm0.34$ & $0.06\pm0.05$ & $0.08\pm0.05$ & $0.53\pm0.08$ & $0.91\pm0.32$ \\
$\mathcal{L}_\text{K}+\mathcal{L}_\text{F}(\text{c})$
& $88\pm2$ & $87\pm2$ & $98\pm1$ & $88\pm2$ 
& $0.05\pm0.08$ & $0.07\pm0.02$ & $0.1\pm0.07$ & $0.41\pm0.05$ & $0.33\pm0.49$ \\
$\mathcal{L}_\text{K}+\mathcal{L}_\text{V}+\mathcal{L}_\text{F}(\text{c})$
& $83\pm1$ & $84\pm2$ & $95\pm1$ & $89\pm3$ 
& $\mathbf{0}\pm0$ & $0.04\pm0.03$ & $0.06\pm0.06$ & $0.5\pm0.16$ & $0.06\pm0.06$ \\
\cmidrule(lr){1-1}\cmidrule(lr){2-5}\cmidrule(lr){6-10}
Roach
& $95\pm5$ & $95\pm1$ & $100\pm0$ & $95\pm1$ 
& $0\pm0$ & $0.04\pm0.04$ & $0.03\pm0.04$ & $0.13\pm0.11$ & $0\pm0$ \\
Autopilot
& $97\pm2$ & $87\pm4$ & $99\pm2$ & $88\pm3$ 
& $0\pm0$ & $0\pm0$ & $0.33\pm0.55$ & $0.89\pm0.54$ & $0.35\pm0.58$ \\
\bottomrule
\end{tabular}
\caption{\textbf{Performance and infraction analysis on NoCrash-busy, train town \& train weather.} Mean and std. over 3 seeds.}
\label{table:infraction_eu_tt}
\end{table*}
\begin{table*}
\setlength{\tabcolsep}{2.6pt}
\centering
\begin{tabular}{lccccccccc} 
\toprule
& \begin{tabular}{@{}c@{}}Success \\ rate \end{tabular} 
& \begin{tabular}{@{}c@{}}Driving \\ score \end{tabular} 
& \begin{tabular}{@{}c@{}}Route \\ compl. \end{tabular} 
& \begin{tabular}{@{}c@{}}Infrac. \\ penalty \end{tabular} 
& \begin{tabular}{@{}c@{}}Collision \\ others \end{tabular} 
& \begin{tabular}{@{}c@{}}Collision \\ pedestrian \end{tabular} 
& \begin{tabular}{@{}c@{}}Collision \\ vehicle \end{tabular}  
& \begin{tabular}{@{}c@{}}Red light \\ infraction \end{tabular}  
& \begin{tabular}{@{}c@{}}Agent \\ blocked \end{tabular}  \\
\cmidrule(lr){1-1}\cmidrule(lr){2-5}\cmidrule(lr){6-10}
iter 5
& \%, $\uparrow$
& \%, $\uparrow$
& \%, $\uparrow$
& \%, $\uparrow$
& \#/Km, $\downarrow$
& \#/Km, $\downarrow$
& \#/Km, $\downarrow$
& \#/Km, $\downarrow$
& \#/Km, $\downarrow$
\\
\cmidrule(lr){1-1}\cmidrule(lr){2-5}\cmidrule(lr){6-10}
$\mathcal{L}_\text{A}(\text{AP})$
& $31\pm3$ & $53\pm2$ & $61\pm1$ & $87\pm1$ 
& $\mathbf{0}\pm0$ & $\mathbf{0}\pm0$ & $0.35\pm0.23$ & $1.31\pm0.36$ & $5.75\pm0.11$ \\
$\mathcal{L}_\text{A}$
& $75\pm4$ & $79\pm5$ & $92\pm2$ & $87\pm4$
& $0.13\pm0.18$ & $0.03\pm0$ & $0.06\pm0.01$ & $0.69\pm0.19$ & $0.79\pm0.32$ \\
$\mathcal{L}_\text{K}$
& $73\pm5$ & $79\pm5$ & $91\pm3$ & $87\pm3$ 
& $0.02\pm0.04$ & $\mathbf{0}\pm0$ & $0.24\pm0.37$ & $0.6\pm0.18$ & $0.95\pm0.45$ \\
$\mathcal{L}_\text{K}+\mathcal{L}_\text{V}$
& $69\pm4$ & $79\pm3$ & $91\pm1$ & $87\pm3$ 
& $0.03\pm0.05$ & $0.04\pm0.03$ & $0.14\pm0.07$ & $0.5\pm0.1$ & $0.7\pm0.05$ \\
$\mathcal{L}_\text{K}+\mathcal{L}_\text{F}$
& $60\pm5$ & $73\pm2$ & $80\pm3$ & $92\pm1$ 
& $0.05\pm0.08$ & $0.1\pm0.16$ & $0.09\pm0.05$ & $0.38\pm0.03$ & $\mathbf{0.02}\pm0.03$ \\
$\mathcal{L}_\text{K}+\mathcal{L}_\text{V}+\mathcal{L}_\text{F}$
& $49\pm8$ & $67\pm4$ & $75\pm4$ & $90\pm1$ 
& $0.07\pm0.13$ & $0.03\pm0.05$ & $0.86\pm1.41$ & $0.88\pm0.61$ & $0.73\pm0.17$ \\
$\mathcal{L}_\text{K}+\mathcal{L}_\text{F}(\text{c})$
& $\mathbf{87}\pm5$ & $\mathbf{90}\pm2$ & $\mathbf{97}\pm2$ & $\mathbf{93}\pm1$ 
& $\mathbf{0}\pm0$ & $0.01\pm0.03$ & $0.03\pm0.06$ & $\mathbf{0.37}\pm0.03$ & $0.23\pm0.13$ \\
$\mathcal{L}_\text{K}+\mathcal{L}_\text{V}+\mathcal{L}_\text{F}(\text{c})$
& $79\pm3$ & $81\pm0$ & $92\pm1$ & $89\pm1$ 
& $\mathbf{0}\pm0$ & $0.01\pm0.01$ & $\mathbf{0.02}\pm0.02$ & $0.57\pm0.06$ & $0.39\pm0.17$  \\
\cmidrule(lr){1-1}\cmidrule(lr){2-5}\cmidrule(lr){6-10}
Roach
& $95\pm5$ & $95\pm1$ & $100\pm0$ & $95\pm1$ 
& $0\pm0$ & $0.04\pm0.04$ & $0.03\pm0.04$ & $0.13\pm0.11$ & $0\pm0$ \\
Autopilot
& $97\pm2$ & $87\pm4$ & $99\pm2$ & $88\pm3$ 
& $0\pm0$ & $0\pm0$ & $0.33\pm0.55$ & $0.89\pm0.54$ & $0.35\pm0.58$ \\
\bottomrule
\end{tabular}
\caption{\textbf{Performance and infraction analysis on NoCrash-busy, train town \& new weather.} Mean and std. over 3 seeds.}
\label{table:infraction_eu_tn}
\end{table*}
\begin{table*}
\setlength{\tabcolsep}{2.4pt}
\centering
\begin{tabular}{lccccccccc} 
\toprule
& \begin{tabular}{@{}c@{}}Success \\ rate \end{tabular} 
& \begin{tabular}{@{}c@{}}Driving \\ score \end{tabular} 
& \begin{tabular}{@{}c@{}}Route \\ compl. \end{tabular} 
& \begin{tabular}{@{}c@{}}Infrac. \\ penalty \end{tabular} 
& \begin{tabular}{@{}c@{}}Collision \\ others \end{tabular} 
& \begin{tabular}{@{}c@{}}Collision \\ pedestrian \end{tabular} 
& \begin{tabular}{@{}c@{}}Collision \\ vehicle \end{tabular}  
& \begin{tabular}{@{}c@{}}Red light \\ infraction \end{tabular}  
& \begin{tabular}{@{}c@{}}Agent \\ blocked \end{tabular}  \\
\cmidrule(lr){1-1}\cmidrule(lr){2-5}\cmidrule(lr){6-10}
iter 5
& \%, $\uparrow$
& \%, $\uparrow$
& \%, $\uparrow$
& \%, $\uparrow$
& \#/Km, $\downarrow$
& \#/Km, $\downarrow$
& \#/Km, $\downarrow$
& \#/Km, $\downarrow$
& \#/Km, $\downarrow$
\\
\cmidrule(lr){1-1}\cmidrule(lr){2-5}\cmidrule(lr){6-10}
$\mathcal{L}_\text{A}(\text{AP})$
& $50\pm5$ & $54\pm1$ & $79\pm3$ & $72\pm3$ 
& $0.88\pm0.86$ & $\mathbf{0}\pm0$ & $\mathbf{0.08}\pm0.06$ & $3.24\pm0.35$ & $3.76\pm0.8$ \\
$\mathcal{L}_\text{A}$
& $73\pm4$ & $81\pm4$ & $94\pm4$ & $85\pm2$
& $1.03\pm1.09$ & $0.09\pm0.05$ & $0.72\pm0.8$ & $0.79\pm0.12$ & $1.24\pm0.88$ \\
$\mathcal{L}_\text{K}$
& $84\pm7$ & $85\pm4$ & $97\pm1$ & $88\pm4$ 
& $0.25\pm0.13$ & $0.02\pm0.03$ & $0.3\pm0.31$ & $0.74\pm0.18$ & $0.37\pm0.04$ \\
$\mathcal{L}_\text{K}+\mathcal{L}_\text{V}$
& $77\pm10$ & $84\pm5$ & $97\pm3$ & $86\pm3$
& $0.25\pm0.28$ & $0.02\pm0.03$ & $0.49\pm0.13$ & $0.73\pm0.18$ & $0.19\pm0.24$ \\
$\mathcal{L}_\text{K}+\mathcal{L}_\text{F}$
& $65\pm2$ & $79\pm2$ & $88\pm1$ & $90\pm3$ 
& $0.31\pm0.47$ & $0.07\pm0.07$ & $0.37\pm0.16$ & $0.6\pm0.19$ & $0.3\pm0.45$ \\
$\mathcal{L}_\text{K}+\mathcal{L}_\text{V}+\mathcal{L}_\text{F}$
& $57\pm4$ & $74\pm4$ & $82\pm1$ & $90\pm4$ 
& $0.96\pm0.2$ & $0.04\pm0.05$ & $0.22\pm0.16$ & $\mathbf{0.43}\pm0.21$ & $0.93\pm0.23$ \\
$\mathcal{L}_\text{K}+\mathcal{L}_\text{F}(\text{c})$
& $89\pm5$ & $\mathbf{90}\pm3$ & $\mathbf{100}\pm1$ & $\mathbf{90}\pm2$
& $\mathbf{0.02}\pm0.03$ & $0.08\pm0.07$ & $0.23\pm0.11$ & $0.59\pm0.12$ & $\mathbf{0.04}\pm0.08$ \\
$\mathcal{L}_\text{K}+\mathcal{L}_\text{V}+\mathcal{L}_\text{F}(\text{c})$
& $\mathbf{91}\pm5$ & $88\pm4$ & $98\pm1$ & $89\pm3$ 
& $0.06\pm0.06$ & $0.01\pm0.03$ & $0.19\pm0.08$ & $0.78\pm0.25$ & $0.06\pm0.06$  \\
\cmidrule(lr){1-1}\cmidrule(lr){2-5}\cmidrule(lr){6-10}
Roach
& $95\pm2$ & $96\pm3$ & $100\pm0$ & $96\pm3$ 
& $0\pm0$ & $0.11\pm0.07$ & $0.04\pm0.05$ & $0.16\pm0.2$ & $0\pm0$ \\
Autopilot
& $91\pm1$ & $79\pm2$ & $98\pm1$ & $80\pm2$ 
& $0\pm0$ & $0\pm0$ & $0.18\pm0.08$ & $1.93\pm0.23$ & $0.18\pm0.08$ \\
\bottomrule
\end{tabular}
\caption{\textbf{Performance and infraction analysis on NoCrash-busy, new town \& train weather.} Mean and std. over 3 seeds.}
\label{table:infraction_eu_nt}
\end{table*}
\begin{table*}
\setlength{\tabcolsep}{3pt}
\centering
\begin{tabular}{lccccccccc} 
\toprule
& \begin{tabular}{@{}c@{}}Success \\ rate \end{tabular} 
& \begin{tabular}{@{}c@{}}Driving \\ score \end{tabular} 
& \begin{tabular}{@{}c@{}}Route \\ compl. \end{tabular} 
& \begin{tabular}{@{}c@{}}Infrac. \\ penalty \end{tabular} 
& \begin{tabular}{@{}c@{}}Collision \\ others \end{tabular} 
& \begin{tabular}{@{}c@{}}Collision \\ pedestrian \end{tabular} 
& \begin{tabular}{@{}c@{}}Collision \\ vehicle \end{tabular}  
& \begin{tabular}{@{}c@{}}Red light \\ infraction \end{tabular}  
& \begin{tabular}{@{}c@{}}Agent \\ blocked \end{tabular}  \\
\cmidrule(lr){1-1}\cmidrule(lr){2-5}\cmidrule(lr){6-10}
iter 5
& \%, $\uparrow$
& \%, $\uparrow$
& \%, $\uparrow$
& \%, $\uparrow$
& \#/Km, $\downarrow$
& \#/Km, $\downarrow$
& \#/Km, $\downarrow$
& \#/Km, $\downarrow$
& \#/Km, $\downarrow$
\\
\cmidrule(lr){1-1}\cmidrule(lr){2-5}\cmidrule(lr){6-10}
$\mathcal{L}_\text{A}(\text{AP})$
& $31 \pm 7$ & $43 \pm 2$ & $62 \pm 6$ & $77 \pm 4$ 
& $0.54 \pm 0.53$ & $\mathbf{0}\pm0$ & $0.63 \pm 0.50$ & $3.33 \pm 0.58$ & $19.4\pm 14.4$ \\
$\mathcal{L}_\text{A}$
& $57\pm7$ & $66\pm3$ & $84\pm3$ & $76\pm1$ 
& $2.07\pm1.37$ & $\mathbf{0}\pm0$ & $1.36\pm1.10$ & $1.4\pm0.2$ & $2.82\pm1.45$ \\
$\mathcal{L}_\text{K}$
& $74\pm3$ & $79\pm0$ & $91\pm2$ & $86\pm1$ 
& $0.50\pm0.25$ & $\mathbf{0}\pm0$ & $0.53\pm0.18$ & $0.68\pm0.08$ & $3.39\pm0.20$ \\
$\mathcal{L}_\text{K}+\mathcal{L}_\text{V}$
& $71\pm9$ & $78\pm3$ & $91\pm1$ & $85\pm3$ 
& $0.55\pm0.22$ & $0.11\pm0.06$ & $0.34\pm0.31$ & $0.72\pm0.09$ & $1.14\pm0.10$ \\
$\mathcal{L}_\text{K}+\mathcal{L}_\text{F}$
& $62\pm2$ & $75\pm1$ & $85\pm0$ & $87\pm2$ 
& $0.79\pm0.61$ & $0.03\pm0.05$ & $0.73\pm0.16$ & $0.63\pm0.02$ & $2.04\pm1.33$ \\
$\mathcal{L}_\text{K}+\mathcal{L}_\text{V}+\mathcal{L}_\text{F}$
& $47\pm9$ & $64\pm6$ & $72\pm5$ & $89\pm3$ 
& $0.9\pm0.73$ & $0.03\pm0.06$ & $0.38\pm0.26$ & $0.79\pm0.42$ & $1.29\pm0.9$ \\
$\mathcal{L}_\text{K}+\mathcal{L}_\text{F}(\text{c})$
& $\mathbf{87} \pm 5$ & $\mathbf{88} \pm 3$ & $\mathbf{96} \pm 0$ & $\mathbf{91} \pm 3$ 
& $\mathbf{0.08} \pm 0.04$ & $0.01 \pm 0.02$ & $0.23 \pm 0.08$ & $\mathbf{0.61} \pm 0.23$ & $0.84 \pm 0.04$ \\
$\mathcal{L}_\text{K}+\mathcal{L}_\text{V}+\mathcal{L}_\text{F}(\text{c})$
& $78\pm3$ & $83\pm1$ & $94\pm2$ & $89\pm2$ 
& $0.21\pm0.14$ & $\mathbf{0}\pm0$ & $\mathbf{0.16}\pm0.05$ & $0.79\pm0.15$ & $\mathbf{0.46}\pm0.13$  \\
\cmidrule(lr){1-1}\cmidrule(lr){2-5}\cmidrule(lr){6-10}
Roach
& $95 \pm 2$ & $96 \pm 3$ & $100 \pm 0$ & $96 \pm 3$ 
& $0 \pm 0$ & $0.11 \pm 0.07$ & $0.04 \pm 0.05$ & $0.16 \pm 0.20$ & $0 \pm 0$ \\
Autopilot
& $91 \pm 1$ & $79 \pm 2$ & $98 \pm 1$ & $80 \pm 2$ 
& $0 \pm 0$ & $0 \pm 0$ & $0.18 \pm 0.08$ & $1.93 \pm 0.23$ & $0.18 \pm 0.08$\\
\bottomrule
\end{tabular}
\caption{\textbf{Performance and infraction analysis on NoCrash-busy, new town \& new weather.} Mean and std. over 3 seeds.}
\label{table:infraction_eu_nn}
\end{table*}

\begin{table*}
\setlength{\tabcolsep}{1.2pt}
\centering
\begin{tabular}{lcccccccccc} 
\toprule
& \begin{tabular}{@{}c@{}}Success \\ rate \end{tabular} 
& \begin{tabular}{@{}c@{}}Driving \\ score \end{tabular} 
& \begin{tabular}{@{}c@{}}Route \\ compl. \end{tabular} 
& \begin{tabular}{@{}c@{}}Infrac. \\ penalty \end{tabular} 
& \begin{tabular}{@{}c@{}}Collision \\ others \end{tabular} 
& \begin{tabular}{@{}c@{}}Collision \\ pedestrian \end{tabular} 
& \begin{tabular}{@{}c@{}}Collision \\ vehicle \end{tabular}  
& \begin{tabular}{@{}c@{}}Red light \\ infraction \end{tabular}  
& \begin{tabular}{@{}c@{}}Stop Sign \\ infraction \end{tabular}  
& \begin{tabular}{@{}c@{}}Agent \\ blocked \end{tabular}  \\
\cmidrule(lr){1-1}\cmidrule(lr){2-5}\cmidrule(lr){6-11}
iter 5
& \%, $\uparrow$
& \%, $\uparrow$
& \%, $\uparrow$
& \%, $\uparrow$
& \#/Km, $\downarrow$
& \#/Km, $\downarrow$
& \#/Km, $\downarrow$
& \#/Km, $\downarrow$
& \#/Km, $\downarrow$
& \#/Km, $\downarrow$
\\
\cmidrule(lr){1-1}\cmidrule(lr){2-5}\cmidrule(lr){6-11}
$\mathcal{L}_\text{A}(\text{AP})$
& $50$ & $55$ & $82$ & $68$ 
& $0.24$ & $\mathbf{0.01}$ & $0.38$ & $0.53$ & $\mathbf{0.22}$ & $1.39$  \\
$\mathcal{L}_\text{A}$
& $51$ & $54$ & $87$ & $60$ 
& $0.46$ & $0.19$ & $0.30$ & $0.50$ & $0.39$ & $0.48$ \\
$\mathcal{L}_\text{K}$
& $44$ & $53$ & $86$ &  $63$
& $0.14$ & $0.07$ & $0.35$ & $0.42$ & $0.38$ & $0.77$  \\
$\mathcal{L}_\text{K}+\mathcal{L}_\text{V}$
& $49$ & $53$ & $81$ & $66$ 
& $0.39$ & $0.04 $ & $0.30 $ & $0.36 $ & $0.40 $ & $1.35 $  \\
$\mathcal{L}_\text{K}+\mathcal{L}_\text{F}$
& $53$ & $60$ & $85$ & $\mathbf{71}$ 
& $0.11 $ & $0.10 $ & $0.20 $ & $\mathbf{0.25} $ & $0.32 $ & $0.47 $ \\
$\mathcal{L}_\text{K}+\mathcal{L}_\text{V}+\mathcal{L}_\text{F}$
& $62$ & $61$ & $94$ & $65$
& $\mathbf{0.01} $ & $0.05 $ & $0.30 $ & $0.37 $ & $0.42 $ & $0.47 $ \\
$\mathcal{L}_\text{K}+\mathcal{L}_\text{F}(\text{c})$
& $\mathbf{69}$ & $\mathbf{62}$ & $94$ & $66$ 
& $0.05$ & $0.04$ & $\mathbf{0.15}$ & $0.35$ & $0.59$ & $\mathbf{0.40}$  \\
$\mathcal{L}_\text{K}+\mathcal{L}_\text{V}+\mathcal{L}_\text{F}(\text{c})$
& $62$ & $59$ & $\mathbf{95}$ & $63$ 
& $0.04$ & $0.41$ & $0.21$ & $0.33$ & $0.50$ & $0.45$  \\
\cmidrule(lr){1-1}\cmidrule(lr){2-5}\cmidrule(lr){6-11}
Roach
& $74\pm1$ & $82\pm2$ & $95\pm1$ & $86\pm2$ 
& $0.03\pm0.02$ & $0.04\pm0.03$ & $0.12\pm0.04$ & $0.13\pm0.05$ & $0\pm0.01$ & $0.13\pm0.04$  \\
Autopilot
& $76\pm1$ & $80\pm1$ & $96\pm1$ & $84\pm2$ 
& $0\pm0$ & $0\pm0$ & $0.16\pm0.05$ & $0.3\pm0.05$ & $0\pm0.01$ & $0.16\pm0.07$  \\
\bottomrule
\end{tabular}
\caption{\textbf{Performance and infraction analysis on the offline LeaderBoard, train town \& train weather.} Mean and std. over 3 seeds.}
\label{table:infraction_lb_tt}
\end{table*}
\begin{table*}
\setlength{\tabcolsep}{1.2pt}
\centering
\begin{tabular}{lcccccccccc} 
\toprule
& \begin{tabular}{@{}c@{}}Success \\ rate \end{tabular} 
& \begin{tabular}{@{}c@{}}Driving \\ score \end{tabular} 
& \begin{tabular}{@{}c@{}}Route \\ compl. \end{tabular} 
& \begin{tabular}{@{}c@{}}Infrac. \\ penalty \end{tabular} 
& \begin{tabular}{@{}c@{}}Collision \\ others \end{tabular} 
& \begin{tabular}{@{}c@{}}Collision \\ pedestrian \end{tabular} 
& \begin{tabular}{@{}c@{}}Collision \\ vehicle \end{tabular}  
& \begin{tabular}{@{}c@{}}Red light \\ infraction \end{tabular}  
& \begin{tabular}{@{}c@{}}Stop Sign \\ infraction \end{tabular}  
& \begin{tabular}{@{}c@{}}Agent \\ blocked \end{tabular}  \\
\cmidrule(lr){1-1}\cmidrule(lr){2-5}\cmidrule(lr){6-11}
iter 5
& \%, $\uparrow$
& \%, $\uparrow$
& \%, $\uparrow$
& \%, $\uparrow$
& \#/Km, $\downarrow$
& \#/Km, $\downarrow$
& \#/Km, $\downarrow$
& \#/Km, $\downarrow$
& \#/Km, $\downarrow$
& \#/Km, $\downarrow$
\\
\cmidrule(lr){1-1}\cmidrule(lr){2-5}\cmidrule(lr){6-11}
$\mathcal{L}_\text{A}(\text{AP})$
& $ 14$ & $ 32$ & $ 47$ & $ \mathbf{79}$ 
& $ 0.23$ & $ \mathbf{0.00}$ & $ 0.31$ & $ 0.55$ & $ 0.32$ & $ 31.79$ \\
$\mathcal{L}_\text{A}$
& $ 55$ & $ 55$ & $ 87$ & $ 64$ 
& $0.14$ & $ 0.03$ & $ 0.26$ & $ 0.37$ & $ 0.47$ & $ 0.43$ \\
$\mathcal{L}_\text{K}$
& $ 50$ & $ 50$ & $ 87$ & $ 58$ 
& $ 0.08$ & $0.06 $ & $0.42 $ & $0.57 $ & $0.62 $ & $0.61 $ \\
$\mathcal{L}_\text{K}+\mathcal{L}_\text{V}$
& $ 40$ & $ 48$ & $ 79$ & $ 64$ 
& $ 0.13$ & $0.02 $ & $0.37 $ & $0.48 $ & $0.45 $ & $0.80 $ \\
$\mathcal{L}_\text{K}+\mathcal{L}_\text{F}$
& $ 43$ & $ 56$ & $ 82$ & $ 70$ 
& $ 0.11$ & $0.03 $ & $0.20 $ & $0.34 $ & $\mathbf{0.31} $ & $0.66 $ \\
$\mathcal{L}_\text{K}+\mathcal{L}_\text{V}+\mathcal{L}_\text{F}$
& $ \mathbf{56}$ & $ \mathbf{62}$ & $ \mathbf{91}$ & $ 69$ 
& $ \mathbf{0.06}$ & $0.05 $ & $\mathbf{0.15} $ & $\mathbf{0.29} $ & $0.39 $ & $\mathbf{0.31} $ \\
$\mathcal{L}_\text{K}+\mathcal{L}_\text{F}(\text{c})$
& $\mathbf{56}$ & $58$ & $90$ & $66$ 
& $0.07$ & $0.05$ & $0.19$ & $0.36$ & $0.60$ & $0.36$  \\
$\mathcal{L}_\text{K}+\mathcal{L}_\text{V}+\mathcal{L}_\text{F}(\text{c})$
& $39$ & $51$ & $88$ & $59$ 
& $0.10$ & $0.03$ & $0.29$ & $0.38$ & $0.54$ & $0.47$  \\
\cmidrule(lr){1-1}\cmidrule(lr){2-5}\cmidrule(lr){6-11}
Roach
& $71\pm2$ & $81\pm1$ & $95\pm1$ & $85\pm0$ 
& $0.02\pm0.02$ & $0.04\pm0.03$ & $0.14\pm0.02$ & $0.12\pm0.04$ & $0\pm0.01$ & $0.14\pm0.07$  \\
Autopilot
& $77\pm2$ & $81\pm1$ & $96\pm1$ & $85\pm2$ 
& $0\pm0$ & $0\pm0$ & $0.16\pm0.04$ & $0.28\pm0.06$ & $0\pm0.01$ & $0.22\pm0.13$  \\
\bottomrule
\end{tabular}
\caption{\textbf{Performance and infraction analysis on the offline LeaderBoard, train town \& new weather.} Mean and std. over 3 seeds.}
\label{table:infraction_lb_tn}
\end{table*}
\begin{table*}
\setlength{\tabcolsep}{1.7pt}
\centering
\begin{tabular}{lcccccccccc} 
\toprule
& \begin{tabular}{@{}c@{}}Success \\ rate \end{tabular} 
& \begin{tabular}{@{}c@{}}Driving \\ score \end{tabular} 
& \begin{tabular}{@{}c@{}}Route \\ compl. \end{tabular} 
& \begin{tabular}{@{}c@{}}Infrac. \\ penalty \end{tabular} 
& \begin{tabular}{@{}c@{}}Collision \\ others \end{tabular} 
& \begin{tabular}{@{}c@{}}Collision \\ pedestrian \end{tabular} 
& \begin{tabular}{@{}c@{}}Collision \\ vehicle \end{tabular}  
& \begin{tabular}{@{}c@{}}Red light \\ infraction \end{tabular}  
& \begin{tabular}{@{}c@{}}Stop Sign \\ infraction \end{tabular}  
& \begin{tabular}{@{}c@{}}Agent \\ blocked \end{tabular}  \\
\cmidrule(lr){1-1}\cmidrule(lr){2-5}\cmidrule(lr){6-11}
iter 5
& \%, $\uparrow$
& \%, $\uparrow$
& \%, $\uparrow$
& \%, $\uparrow$
& \#/Km, $\downarrow$
& \#/Km, $\downarrow$
& \#/Km, $\downarrow$
& \#/Km, $\downarrow$
& \#/Km, $\downarrow$
& \#/Km, $\downarrow$
\\
\cmidrule(lr){1-1}\cmidrule(lr){2-5}\cmidrule(lr){6-11}
$\mathcal{L}_\text{A}(\text{AP})$
& $ 35$ & $ 37$ & $ 75$ & $ 55$ 
& $ 0.17$ & $ \mathbf{0.00}$ & $ 1.52$ & $ 1.00$ & $ 0.50$ & $ 3.64$ \\
$\mathcal{L}_\text{A}$
& $ 58$ & $ 42$ & $ 92$ & $ 46$ 
& $ 0.17$ & $ 0.04$ & $ 0.42$ & $ 0.82$ & $ 0.75$ & $ 0.29$ \\
$\mathcal{L}_\text{K}$
& $ 44$ & $ 40$ & $ 91$ & $ 44$ 
& $ 0.91$ & $ 0.04$ & $ 0.36$ & $ 1.21$ & $ 0.75$ & $ 0.94$ \\
$\mathcal{L}_\text{K}+\mathcal{L}_\text{V}$
& $ 37$ & $ 40$ & $ 76$ & $ \mathbf{58}$ 
& $ 0.13$ & $ 0.05$ & $ 0.45$ & $ 0.80$ & $ \mathbf{0.40}$ & $ 0.33$ \\
$\mathcal{L}_\text{K}+\mathcal{L}_\text{F}$
& $ 56$ & $ 51$ & $ 91$ & $ 56$ 
& $ 0.22$ & $0.07 $ & $0.34$ & $0.45 $ & $0.61 $ & $0.16$ \\
$\mathcal{L}_\text{K}+\mathcal{L}_\text{V}+\mathcal{L}_\text{F}$
& $ 60$ & $ 47$ & $ \mathbf{95}$ & $ 50$ 
& $ \mathbf{0.00}$ & $0.09 $ & $0.81 $ & $0.64 $ & $0.74 $ & $1.29 $ \\
$\mathcal{L}_\text{K}+\mathcal{L}_\text{F}(\text{c})$
& $62$ & $53$ & $94$ & $56$ 
& $0.00$ & $0.04$ & $1.22$ & $0.71$ & $0.70$ & $1.04$  \\
$\mathcal{L}_\text{K}+\mathcal{L}_\text{V}+\mathcal{L}_\text{F}(\text{c})$
& $\mathbf{67}$ & $\mathbf{56}$ & $\mathbf{95}$ & $\mathbf{58}$ 
& $0.03$ & $0.05$ & $\mathbf{0.31}$ & $\mathbf{0.38}$ & $0.72$ & $\mathbf{0.11}$  \\
\cmidrule(lr){1-1}\cmidrule(lr){2-5}\cmidrule(lr){6-11}
Roach
& $78\pm4$ & $83\pm2$ & $97\pm1$ & $86\pm2$ 
& $0\pm0$ & $0.03\pm0.02$ & $0.13\pm0.1$ & $0.16\pm0.03$ & $0\pm0$ & $0.09\pm0.04$  \\
Autopilot
& $72\pm13$ & $74\pm5$ & $95\pm2$ & $78\pm3$ 
& $0\pm0$ & $0\pm0$ & $0.14\pm0.07$ & $0.57\pm0.13$ & $0\pm0$ & $0.18\pm0.14$  \\
\bottomrule
\end{tabular}
\caption{\textbf{Performance and infraction analysis on the offline LeaderBoard, new town \& train weather.} Mean and std. over 3 seeds.}
\label{table:infraction_lb_nt}
\end{table*}
\begin{table*}
\setlength{\tabcolsep}{2pt}
\centering
\begin{tabular}{lcccccccccc} 
\toprule
& \begin{tabular}{@{}c@{}}Success \\ rate \end{tabular} 
& \begin{tabular}{@{}c@{}}Driving \\ score \end{tabular} 
& \begin{tabular}{@{}c@{}}Route \\ compl. \end{tabular} 
& \begin{tabular}{@{}c@{}}Infrac. \\ penalty \end{tabular} 
& \begin{tabular}{@{}c@{}}Collision \\ others \end{tabular} 
& \begin{tabular}{@{}c@{}}Collision \\ pedestrian \end{tabular} 
& \begin{tabular}{@{}c@{}}Collision \\ vehicle \end{tabular}  
& \begin{tabular}{@{}c@{}}Red light \\ infraction \end{tabular}  
& \begin{tabular}{@{}c@{}}Stop Sign \\ infraction \end{tabular}  
& \begin{tabular}{@{}c@{}}Agent \\ blocked \end{tabular}  \\
\cmidrule(lr){1-1}\cmidrule(lr){2-5}\cmidrule(lr){6-11}
iter 5
& \%, $\uparrow$
& \%, $\uparrow$
& \%, $\uparrow$
& \%, $\uparrow$
& \#/Km, $\downarrow$
& \#/Km, $\downarrow$
& \#/Km, $\downarrow$
& \#/Km, $\downarrow$
& \#/Km, $\downarrow$
& \#/Km, $\downarrow$
\\
\cmidrule(lr){1-1}\cmidrule(lr){2-5}\cmidrule(lr){6-11}
$\mathcal{L}_\text{A}(\text{AP})$
& $ 14$ & $ 30$ & $ 42$ & $ \mathbf{80}$ 
& $ 0.06$ & $0.05 $ & $1.11 $ & $0.63 $ & $\mathbf{0.49} $ & $28.27 $ \\
$\mathcal{L}_\text{A}$
& $40$ & $ 41$ & $ \mathbf{90}$ & $ 45$ 
& $ 0.36$ & $ 0.10$ & $0.61 $ & $0.88 $ & $0.67 $ & $\mathbf{0.35} $ \\
$\mathcal{L}_\text{K}$
& $ 39$ & $ 30$ & $ 86$ & $ 39$ 
& $ 0.17$ & $ 0.03$ & $0.42$ & $ 1.31$ & $ 0.81$ & $ 0.51$ \\
$\mathcal{L}_\text{K}+\mathcal{L}_\text{V}$
& $ 33$ & $ 37$ & $ 78$ & $ 53$ 
& $ 0.09$ & $ 0.06$ & $ 0.47$ & $ 0.97$ & $ 0.54$ & $ 0.40$ \\
$\mathcal{L}_\text{K}+\mathcal{L}_\text{F}$
& $ 37$ & $ 41$ & $ 81$ & $ 54$ 
& $ 0.29$ & $ 0.03$ & $ 0.79$ & $0.61$ & $ 0.68$ & $ 1.23$ \\
$\mathcal{L}_\text{K}+\mathcal{L}_\text{V}+\mathcal{L}_\text{F}$
& $ 39$ & $ 45$ & $ 85$ & $ 56$ 
& $0.02$ & $ \mathbf{0.01}$ & $ 1.54$ & $ 0.73$ & $ 0.64$ & $ 2.30$ \\
$\mathcal{L}_\text{K}+\mathcal{L}_\text{F}(\text{c})$
& $\mathbf{50}$ & $\mathbf{50}$ & $86$ & $60$ 
& $\mathbf{0.01}$ & $0.02$ & $0.48$ & $\mathbf{0.60}$ & $0.63$ & $2.64$  \\
$\mathcal{L}_\text{K}+\mathcal{L}_\text{V}+\mathcal{L}_\text{F}(\text{c})$
& $48$ & $48$ & $\mathbf{90}$ & $56$ 
& $0.02$ & $0.04$ & $\mathbf{0.18}$ & $\mathbf{0.60}$ & $0.81$ & $0.47$  \\
\cmidrule(lr){1-1}\cmidrule(lr){2-5}\cmidrule(lr){6-11}
Roach
& $78\pm4$ & $83\pm2$ & $97\pm1$ & $85\pm2$ 
& $0\pm0$ & $0.04\pm0.02$ & $0.13\pm0.1$ & $0.18\pm0.06$ & $0\pm0$ & $0.09\pm0.04$  \\
Autopilot
& $71\pm11$ & $74\pm4$ & $95\pm1$ & $78\pm3$ 
& $0\pm0$ & $0\pm0$ & $0.14\pm0.07$ & $0.58\pm0.12$ & $0\pm0$ & $0.2\pm0.12$  \\
\bottomrule
\end{tabular}
\caption{\textbf{Performance and infraction analysis on the offline LeaderBoard, new town \& new weather.} Mean and std. over 3 seeds.}
\label{table:infraction_lb_nn}
\end{table*}

\end{document}